\documentclass[11pt]{article}

\PassOptionsToPackage{table}{xcolor}
\usepackage[preprint]{acl}
\usepackage{times}
\usepackage{latexsym}
\usepackage[T1]{fontenc}
\usepackage[utf8]{inputenc}
\usepackage{microtype}
\usepackage{inconsolata}
\usepackage{graphicx}
\usepackage{amsmath}
\usepackage{amssymb}
\usepackage{amsthm}
\usepackage{booktabs}
\usepackage{xcolor}
\usepackage{enumitem}
\usepackage[capitalize,noabbrev]{cleveref}

\usepackage{tabularx}
\usepackage{array}

\usepackage[breakable]{tcolorbox}

\newtcolorbox{definitionbox}[1]{
    breakable,
    colback=black!4,
    colframe=black!25,
    colbacktitle=black!12,
    coltitle=black,
    fonttitle=\bfseries\small,
    title={#1},
    toptitle=2pt,
    bottomtitle=2pt,
    boxrule=0.4pt,
    arc=0pt,
    boxsep=0pt,
    left=6pt,
    right=6pt,
    top=5pt,
    bottom=5pt,
    before skip=8pt,
    after skip=8pt
}

\usepackage{float}
\newfloat{termsfloat}{t}{lod}

\newcommand{\hd}[2]{\begin{tabular}[b]{@{}c@{}}#1\\#2\end{tabular}}
\newcommand{\B}{B}
\newcommand{\off}{\textsc{vanilla}}
\newcommand{\fo}{\textsc{frontier-only}}
\newcommand{\seq}{\textsc{one-per-step}}
\newcommand{\ltr}{\textsc{left-to-right}}

\newcommand{\eosf}{\textsc{eos-at-frontier}}
\newcommand{\fa}{\textsc{frontier-fallback}}
\newcommand{\gate}{\textsc{frontier-gating}}
\newcommand{\ebud}{\textsc{entropy-budget}}

\title{Fast at the Frontier, Wrong at the Fallback: A Commit-Level Analysis of Threshold Decoding in Diffusion Language Models}

\author{Roman Makarov \\ MBZUAI \\ Abu Dhabi, UAE \\ \texttt{Roman.Makarov@mbzuai.ac.ae} \And Zhe Xu \\ UIUC \\ Urbana, IL, USA \\ \texttt{zhexu3@illinois.edu} \And Jian Kang \\ MBZUAI \\ Abu Dhabi, UAE \\ \texttt{Jian.Kang@mbzuai.ac.ae}}

\begin{document}
\maketitle

\begin{abstract}
Diffusion language models (DLMs) can write several tokens per forward pass. Open-weight DLMs commit, in any order, every masked position whose confidence passes a threshold. If none does, a fallback commits the most confident one, and the commits are never revised. Wide windows let the decoder commit far ahead of the text, which should speed up decoding but often costs accuracy. It is unclear where the model is confident and which commits cause errors. We trace every commit of four off-the-shelf DLMs on math, code, translation and summarization, and rerun failing answers with one step changed. Most tokens become confident only near the frontier, the leftmost masked position, so committing only there, still several tokens per step, is about as fast as committing anywhere. The accuracy loss comes from committing tokens before the text to their left, and committing one token per step does not reduce it. The worst case is an early end-of-sequence token that cuts the answer short. Fallback commits are 11--35\% of commits but start 61--72\% of the failures that left-to-right decoding avoids. Changing only the step where such a failure starts repairs it more often than changing a similar earlier step. Committing only at the frontier recovers most of the lost accuracy on all four models. Our code is available at \url{https://github.com/rmakarovv/fast-at-the-frontier}.

\end{abstract}

\begin{figure*}[t]
 \centering
 \includegraphics[width=0.86\textwidth]{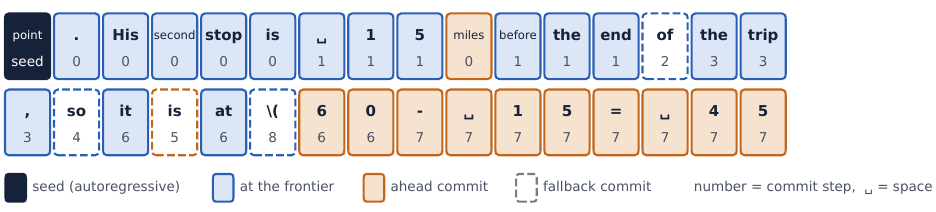}
 \vspace{-0.6em}
 \caption{One real 32-token block (NLD-3B, GSM8K, \off{}); each box shows a token and its commit step. Blue: commits at the frontier; orange: ahead commits, made while a position to their left was still masked (``miles'' at step~0, before ``15''); dashed: fallback commits.}
\label{fig:block}
\end{figure*}

\section{Introduction}
Diffusion language models (DLMs) write an answer by filling in blanks. The answer starts as a row of masked positions. At each step, one forward pass predicts a token for every masked position, and the decoder fills some of them~\citep{sahoo2024mdlm,nie2025llada,arriola2025block}. An autoregressive (AR) model fills one position per pass, from left to right; a DLM can fill several, so it may decode faster. Open-weight DLMs such as LLaDA~\citep{nie2025llada}, Fast-dLLM v2~\citep{wu2025fastdllmv2} and Nemotron-Labs-Diffusion~\citep{fu2026nld} use a confidence-based decoding policy~\citep{wu2025fastdllm}. The decoder works inside a range of answer positions, its \emph{window}. At each step, it fills every masked position in the window whose most likely token has probability above a threshold. If none does, a \emph{fallback} fills the single most confident position. We call each filled token a \emph{commit}. These decoders never change a commit, and they may commit in any order.
A wider window therefore looks like a cheap way to commit more tokens per step.

Following \citet{yeom2026answerfirst}, we call the leftmost masked position the \emph{frontier}. An \emph{ahead commit} fills a position while a position to its left is still masked. For example, ``miles'' in \cref{fig:block} is committed before the ``15'' to its left. Since commits are never revised, an ahead commit is a guess that the policy makes before the context to its left exists. A particularly harmful case is an ahead commit of an end-of-sequence (EOS) token, which forces the answer to finish before the chosen position.

Wide windows can cost a lot of accuracy: on GSM8K, LLaDA scores 77\% with 32-token windows but 14\% with one 256-token window~\citep{liang2026cd4lm}.
Recent policies keep commits near the frontier and recover accuracy. Representative examples include PARD~\citep{hou2026pard}, Fork-dLLM~\citep{frkovic2026fork}, RPD~\citep{he2026rpd}, and frontier gating~\citep{yeom2026answerfirst}. However, by limiting where a DLM can commit, these policies give up the freedom to commit anywhere, which may slow down generation. Whether this freedom is worth keeping decides how DLMs should be decoded and what future models should be trained for. For the standard decoding policy in these models, which we call the \emph{vanilla} policy, it is unclear whether the model becomes confident about masked positions far from the frontier or only about those near it, and which of its commits cause the errors.

Answering these questions requires following single commits, because a bad commit may hurt the answer many tokens later, and averages over all commits, such as tokens per step or how closely the order follows left to right~\citep{zhong2026parallelism}, mix good and bad commits together. We ask three questions:

\begin{itemize}[leftmargin=*, nosep]
    \item \textbf{RQ1 (confidence).} How does the model's confidence change with distance from the frontier, and does committing only at the frontier slow decoding?
    \item \textbf{RQ2 (accuracy).} Why do wide windows lose accuracy, and can a simple frontier policy recover it without slowing decoding?
    \item \textbf{RQ3 (commits).} Which individual commits start the failures?
\end{itemize}
To answer these questions, we trace every commit of four models: Nemotron-Labs-Diffusion (NLD) 3B and 8B, LLaDA-8B and Fast-dLLM v2 7B. We use math and code benchmarks, plus translation and summarization on the two models that lose the most on math. For each commit, we record how far it is from the frontier and whether the fallback made it.

We compare decoding policies that differ in one part only, such as where the fallback commits. Our main test is a \emph{replay}. Once two decodes differ at one step, all later steps differ too, so we take a failing decode, change only the step where it first departs from a left-to-right decode of the same problem, and continue decoding. As a control, we change a similar earlier step of the same decode. If changing the first departure step repairs the answer more often, that step more likely caused the failure. We also use two \emph{diagnostic} policies, each a controlled ablation of \off{} that removes one suspected cause of errors: (1) \eosf{}, which allows the EOS token only at the frontier, and (2) the \emph{distance cap}, which allows commits only within $k$ positions of the frontier (an extension of \citet{yeom2026answerfirst}). We find that:
\begin{itemize}[leftmargin=*, nosep]
    \item \textbf{Most tokens become confident only near the frontier (RQ1).} So a policy that commits only there runs as fast as \off. Commits ahead of the frontier add almost no speed, making wider windows inefficient, especially at large batch sizes (\cref{sec:speed}).
    \item \textbf{The order of commits causes the accuracy loss (RQ2).} Committing one token per step barely helps if the decoder still picks tokens by confidence. The loss comes from early EOS and early content tokens, in a mix that depends on the model, and committing only at the frontier recovers most of it on all four models (\cref{sec:order}).
    \item \textbf{Most failures start when the fallback commits ahead of the frontier (RQ3).} The fallback makes only 11--35\% of all commits, yet these commits start 61--72\% of the failures that left-to-right decoding avoids. When we replay a failing decode and change only that step, the answer is fixed more often than when we change a similar earlier step (\cref{sec:risk}).
\end{itemize}
Decoders should therefore commit only at the frontier, and faster decoding should come from committing more tokens at the frontier in each step rather than from wider windows.

\section{Preliminaries}
\label{sec:prelim}

\paragraph{Masked diffusion language models.}
Masked diffusion language models (MDLMs) generate token sequences by gradually recovering tokens from masked inputs~\citep{austin2021structured,sahoo2024mdlm,shi2024simplified,nie2025llada}. Let $\mathbf{x}=(x_1,\ldots,x_N)\in\mathcal{V}^N$ be a sequence of $N$ tokens from vocabulary $\mathcal{V}$, and let $m$ be a special mask token. The forward diffusion process replaces each token with $m$ independently, according to a predefined noise schedule. With a linear noise schedule, the forward process at time $t\in[0,1]$ is
\begin{equation}
    q_t(z_{t,i}\mid x_i)
    =(1-t)\delta_{x_i}(z_{t,i})
    +t\delta_m(z_{t,i}),
\end{equation}
where $\delta_a$ denotes a point mass at $a$. A denoising model $p_\theta$ learns to recover the original tokens from the corrupted sequence $\mathbf{z}_t$. For the linear schedule, the standard continuous-time training objective reduces to a weighted masked-token prediction loss:
\begin{equation}
\label{eq:mdlm_loss}
    \mathcal{L}
    =
    -\underset{\mathbf{x},t,\mathbf{z}_t}{\mathbb{E}}
    \Big[
    \frac{1}{t}
    \sum_{i=1}^{N}
    \mathbb{I}[z_{t,i}=m]
    \cdot\log p_\theta(x_i\mid\mathbf{z}_t)
    \Big],
\end{equation}
where $t\sim\mathcal{U}(0,1)$ and
$\mathbf{z}_t\sim q_t(\cdot\mid\mathbf{x})$.

\paragraph{Blockwise masked diffusion.}
Blockwise diffusion combines autoregressive dependencies across blocks with parallel denoising within each block \citep{arriola2025block}.
Given a block size $B$, we partition the sequence into $J=\lceil N/B\rceil$ consecutive blocks,
$\mathbf{x}=(\mathbf{x}^{(1)},\ldots,\mathbf{x}^{(J)})$.
The sequence distribution factorizes as
\begin{equation}
\label{eq:block_factorization}
    p_\theta(\mathbf{x})
    =
    \prod_{b=1}^{J}
    p_\theta(\mathbf{x}^{(b)}
    \mid\mathbf{x}^{(<b)}),
\end{equation} where $\mathbf{x}^{(<b)}$ denotes blocks before the $b$-th block. Each block is generated through a masked diffusion process conditioned on the completed prefix. The denoising model predicts masked tokens using bidirectional dependencies within the current block while preserving causal dependencies across blocks. Training follows the masked diffusion objective in Eq.~\eqref{eq:mdlm_loss}, with the unmasked preceding blocks provided as conditioning context.

At inference, the decoder generates blocks from left to right. It starts block $b$ with all positions masked, $\mathbf{z}^{(b)}_0=(m,\ldots,m)$. At each step $s$, the model predicts token distributions for all masked positions in parallel. The decoder selects a subset $\mathcal{C}_s$ to commit, using a fixed rule, such as a confidence threshold, or a learned policy. Let $\hat{x}_{s,i}$ be the predicted token at position $i$. The decoder updates the state as follows:
\begin{equation}
\label{eq:monotone_update}
    z^{(b)}_{s+1,i}
    =
    \begin{cases}
        \hat{x}_{s,i}, & i\in\mathcal{C}_s,\\
        z^{(b)}_{s,i}, & \text{otherwise}.
    \end{cases}
\end{equation}

In this paper, we study \emph{irreversible decoding}: once the decoder commits a token, it does not change it. We evaluate Nemotron-Labs-Diffusion (NLD), LLaDA, and Fast-dLLM v2~\citep{fu2026nld,nie2025llada,wu2025fastdllmv2}. The set of masked positions can only shrink until the block is complete. Each evaluation can commit multiple tokens.

\paragraph{Confidence-based decoding.} The threshold policy is defined as follows: at each step, the decoder commits every masked position in the window whose confidence is at least a threshold (0.9 here; \citealp{wu2025fastdllm}). If none does, the \emph{fallback} commits the most confident one. We call this policy \off{}.

\paragraph{Key terms.} We describe each committed token by where it is relative to the frontier and by whether the fallback made it (\cref{fig:block}). We define these terms below. Positions after answer's first EOS hold EOS \emph{padding}; all other tokens are called \emph{content} tokens.

\begin{termsfloat}[!ht]
\begin{definitionbox}{Terms for commits}
\textbf{Frontier:} the leftmost masked position in the window. Left-to-right decoding commits only the frontier.\\[2pt]
\textbf{Ahead commit:} a commit which has a position to its left that is still masked.\\[2pt]
\textbf{Fallback commit:} a commit made by the fallback when no position reaches the threshold; all other commits are \emph{confident}.
\end{definitionbox}
\end{termsfloat}

\section{Experimental setup}
\label{sec:setup}

\textbf{Models.} NLD 3B/8B~\citep{fu2026nld} are block diffusion models trained with $\B=32$. LLaDA-8B-Instruct~\citep{nie2025llada} is trained without blocks. At inference, we decode its 256-token answer either as one block ($\B=256$) or with $\B=32$. Fast-dLLM v2 7B~\citep{wu2025fastdllmv2} adapts Qwen2.5~\citep{qwen2024qwen25} to block diffusion with $\B=32$, where each forward pass sees the whole block, but the decoder commits only inside the current \emph{sub-block} of 8 tokens. A step commits only inside its \emph{window}: the current block on NLD and LLaDA, and the current sub-block on Fast-dLLM v2. For Fast-dLLM v2 at $\B=64$ and 128, we turn sub-blocks off, so its window is also the whole block. We call the block size used in training ($\B=32$ for NLD and Fast-dLLM v2) the \emph{training block}. Most results use each model's \emph{wide window}: $\B=128$ or 256 for NLD, $\B=256$ (one block) for LLaDA, and $\B=128$ for Fast-dLLM v2.

\begin{table*}[t]
\centering
\caption{\textbf{Decoding policies.} All policies use the same model, prompt, and confidence threshold ($0.9$). Commits are restricted to the decoding window. The first four vary position order and token count. The remaining policies modify \off{}.}
\label{tab:rules}
\footnotesize
\renewcommand{\arraystretch}{1.05}
\setlength{\tabcolsep}{7pt}
\begin{tabularx}{\textwidth}{
    @{}
    l
    >{\raggedright\arraybackslash}X
    @{}
}
\toprule
\textbf{Policy} & \textbf{Tokens committed per step} \\
\midrule

\off{} &
\textbf{One or more.}
Commit all tokens with confidence $\geq 0.9$, otherwise the most confident one.\\

\seq{} &
\textbf{One.}
Commit the most confident masked position. \\

\ltr{} &
\textbf{One.}
Commit the frontier position (left-to-right decoding). \\

\fo{} &
\textbf{One or more.}
Commit the confident positions in a row, starting at the frontier.
If none exists, commit the frontier position (PARD; \citealp{hou2026pard}). \\

\midrule
\multicolumn{2}{@{}l}{
    \textit{Variants of \off{}: \textbf{one or more} tokens per step}
} \\

\fa{} &
Use the frontier position as the fallback (Fork-dLLM; \citealp{frkovic2026fork}). \\

\gate{} &
Limit commits to the first 16 positions from the
frontier~\citep{yeom2026answerfirst}. \\

\ebud{} &
Apply RPD's 4-nat entropy budget to masked positions
on the left~\citep{he2026rpd}. \\

\eosf{} &
Allow the end-of-sequence token only at the frontier
(ours). \\

Cap $k$ &
Limit commits to within $k$ positions of the frontier
(ours, extends \citealp{yeom2026answerfirst}). \\

\bottomrule
\end{tabularx}
\end{table*}

\noindent \textbf{Decoding policies.} \Cref{tab:rules} lists the policies: \off{}, two one-token baselines (\seq{} and \ltr{}), four policies from recent work, and two diagnostic policies, each of which removes one suspected cause of errors.

\noindent \textbf{Tasks.} We use math (GSM8K, \citealp{cobbe2021gsm8k}; MATH-500, \citealp{hendrycks2021math,lightman2023verify}) and code (HumanEval, \citealp{chen2021codex}; MBPP, \citealp{austin2021mbpp}) benchmarks, plus translation (WMT16, \citealp{bojar2016wmt}) and summarization (CNN/DailyMail, \citealp{hermann2015cnndm}) on NLD-3B and LLaDA.

\noindent \textbf{Implementation details.} Main comparisons use greedy decoding. In addition, we sample three seeds at temperature 0.7. The quality of DLM decoders can depend on the prompt~\citep{evalillusion2026}, so we run LLaDA with two prompts (A: step by step; B: the CD4LM evaluation prompt, \citealp{liang2026cd4lm}). For every commit, we log its position, step, confidence, distance and whether the fallback made it (\cref{fig:block}). \emph{Tokens per forward} (TPF) is the number of committed tokens per forward pass. \Cref{app:setup} gives the tasks, the settings of the baseline policies and how we grade answers.

\noindent \textbf{Statistics.} We compare accuracies with paired exact McNemar tests~\citep{mcnemar1947}, Holm-corrected~\citep{holm1979}, and report a difference as its estimate $\pm$ half the width of its 95\% bootstrap interval over problems. Accuracy differences are in percentage points (\emph{points}).

\section{Most tokens become confident only near the frontier}
\label{sec:speed}

\begin{figure*}[t]
 \centering
 \includegraphics[width=\textwidth]{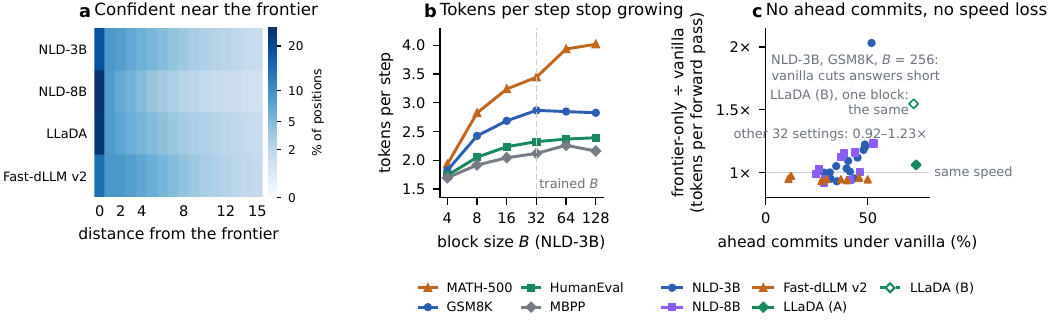}
 \vspace{-1.6em}
 \caption{\textbf{Most tokens become confident only near the frontier, so wider windows add little speed.} \textbf{(a)} How far ahead of the frontier answer positions first pass the threshold in left-to-right decodes (GSM8K; NLD $\B=128$, LLaDA $\B=256$ prompt A, Fast-dLLM v2 $\B=128$), in \% of answer positions per distance. \textbf{(b)} Tokens per step by $\B$ (NLD-3B, \off{}, blocks inside the answer). \textbf{(c)} TPF of \fo{} over TPF of \off{} at the same window, against \off{}'s ahead share; one marker per model, task and window (34 in all). The two high points are settings where \off{} cuts answers short.}
 \label{fig:speed3}
\end{figure*}

Under threshold decoding, a step commits every position whose confidence passes the threshold, so the number of committed tokens depends on how many positions are confident at the same step. To analyze how the model's confidence varies across masked positions (RQ1), we compare \off{} with \fo{}, which commits only at the frontier.

\noindent \textbf{Most tokens become confident only near the frontier.} In left-to-right decodes, for each answer position, we log the first step at which its confidence passes the threshold, and its \textit{distance} to the frontier at that step. Specifically, for a commit at position $j$ and let $f$ be the frontier, we define the distance $d = j - f$.
On all four models, 15--24\% of positions pass the threshold only once they are the frontier, and 60--64\% within seven positions of it. At a distance of 16 positions, the share of positions that pass the threshold for the first time falls below 1.5\% (\cref{fig:speed3}a). The tokens a step can commit are therefore mostly within a few positions of the frontier, and a wider window mainly adds positions that become confident only later.

\noindent \textbf{Larger blocks do not commit more per step.} On NLD-3B, a position at a given offset in its block is committed at about the same step for every $\B$ from 8 to 128, so the number of committed tokens per step stops growing near the training block (\cref{fig:speed3}b). LLaDA and Fast-dLLM v2 behave the same way, and only NLD-8B, and NLD-3B on MATH-500, still gain 10--17\% tokens per step from $\B=32$ to 128. Hence, the speed of threshold decoding is limited by how many tokens are confident near the frontier. \Cref{app:prop} gives the per-position curves.

\noindent \textbf{Removing ahead commits keeps the speed.} With wide windows, about half of \off{}'s commits on NLD and about three quarters on LLaDA are ahead commits (\cref{fig:speed3}c, horizontal axis). Yet \fo{}, which makes no ahead commits, has TPF at least 0.92 times that of \off{} across the 34 settings in \cref{fig:speed3}. In other words, almost all of the speed comes from commits at the frontier, and ahead commits can be dropped without slowing decoding down.

\noindent \textbf{Wider windows cost compute.} Each forward pass processes the whole block, so a wider window adds compute per answer token. On NLD-3B, this is nearly free at batch size 1, but at batch size 32, $\B=128$ is 2.4--2.5 times slower than $\B=32$. Thus, wider windows make each forward pass more expensive, but save few passes, making decoding slower at larger concurrency. \Cref{app:speed} gives TPF and throughput for every setting.

\noindent \textbf{Summary (RQ1).} Most tokens become confident only near the frontier, ahead commits add little speed, and wider windows mainly add compute. Thus, faster decoding should come from committing more tokens at the frontier in each step. NLD's self-speculation does this~\citep{fu2026nld,sahoo2026uno} and is the fastest decoding policy we measured.

\section{The order of commits causes the accuracy loss}
\label{sec:order}

\begin{table*}[t]
\caption{\textbf{Frontier order beats confidence order at every wide window; committing one token per step does not help.} Allowing EOS only at the frontier removes the gap on NLD, but not on LLaDA or Fast-dLLM v2. GSM8K accuracy (\%) at each model's wide window. \emph{Order}: frontier minus confidence order, averaged over one and several tokens per step, $\tfrac12[(\text{\fo{}}-\text{\off{}})+(\text{\ltr{}}-\text{\seq{}})]$; \emph{parallelism}: several minus one token per step, averaged over both orders. \emph{EOS only at the frontier}: the confidence-order policies rerun with this restriction and the order effect recomputed; near zero means that early EOS tokens caused the order effect. Intervals in \cref{app:tables}.}
\label{tab:main}
\centering
\small
\setlength{\tabcolsep}{2.9pt}
\begin{tabular}{llcccccccccc}
\toprule
 & & \multicolumn{2}{c}{\textbf{confidence order}} & \multicolumn{2}{c}{\textbf{frontier order}} & \multicolumn{2}{c}{\textbf{effect (points)}} &
\multicolumn{3}{c}{\textbf{EOS only at the frontier}} \\
\cmidrule(lr){3-4}\cmidrule(lr){5-6}\cmidrule(lr){7-8}\cmidrule(lr){9-11}
\textbf{Model} & $\B$ & \hd{several}{\tiny\off{}} & \hd{one}{\tiny\seq{}} & \hd{one}{\tiny\ltr{}} & \hd{several}{\tiny\fo{}}
& order & \hd{paral-}{lelism} & \hd{several}{\tiny\eosf{}} & \hd{one}{\tiny\seq{}} & order \\
\midrule
NLD-3B & 128 & 79.1 & 79.7 & 88.0 & 87.6 & $+8.4$ & $-0.5$ & 86.2 & 87.3 & $+1.1$ \\
NLD-3B & 256 & 54.5 & 55.0 & 88.0 & 88.1 & $+33.3$ & $-0.2$ & 87.2 & 88.0 & $+0.5$ \\
NLD-8B & 128 & 87.3 & 86.7 & 90.2 & 90.1 & $+3.2$ & $+0.2$ & 91.0 & 91.0 & $-0.8$ \\
NLD-8B & 256 & 73.8 & 73.3 & 88.9 & 88.8 & $+15.3$ & $+0.2$ & 90.4 & 90.2 & $-1.4$ \\
LLaDA, A & 256 & 50.6 & 49.9 & 71.9 & 71.6 & $+21.5$ & $+0.2$ & 54.0 & 53.9 & $+17.8$ \\
LLaDA, B & 256 & 35.5 & 36.2 & 82.2 & 82.3 & $+46.4$ & $-0.3$ & 57.4 & 57.5 & $+24.8$ \\
Fast-dLLM v2 & 128 & 79.4 & 82.0 & 86.6 & 85.4 & $+5.3$ & $-1.9$ & 79.5 & 82.0 & $+5.3$ \\
\bottomrule
\end{tabular}
\end{table*}

\off{} commits several tokens per step and picks them by confidence alone, ignoring their position. Errors could come from either choice: committing tokens in parallel, or ignoring where they are. To find which choice costs accuracy (RQ2), we change one at a time (\cref{tab:main}). If committing in parallel were the problem, policies that commit one token per step would recover the accuracy. If ignoring position were the problem, policies that commit only at the frontier would recover it.

\subsection{Order costs more accuracy than parallelism}
\label{sec:factorial}

On GSM8K, policies that commit tokens only at the frontier (\ltr{}, \fo{}) beat policies that commit confident tokens anywhere in the window (\seq{}, \off{}) on every model by 3 to 46 points, while committing one token per step instead of several changes accuracy by at most two points (\cref{tab:main}). What matters is which tokens the decoder commits: confidence order commits tokens ahead of the frontier, before the text to their left exists. A decoder can therefore keep the accuracy and still commit several tokens per step, as long as it commits them at the frontier. \Cref{app:tables} repeats the comparison on every task; parallelism costs a few points only on MATH-500.

\subsection{Early EOS tokens cut answers short}
\label{sec:eos}

When most of the tokens are still masked in a wide window, often no position is confident enough to pass the threshold. In that case the fallback commits the most confident position anywhere in the window, which is often an EOS token closer to the end of the block. We call such a commit an \textit{early EOS}.
We hypothesize that it happens due to the EOS padding that the model sees during the training. If the fallback commits EOS at position $j$ while the frontier is at position $i < j$, the rest of the answer must fit into $j - i$ positions, which is often too few. On NLD-3B, answers with an early EOS from the fallback are 48\% correct at $\B=128$, and 26\% at 256, while 85--89\% of the same problems are solved with \eosf{} decoding. In smaller windows ($\B\leq64$), EOS usually marks the right length. \Cref{app:eos} gives the full counts.

\noindent \textbf{Two causes, mixed differently by model.} To separate the EOS effect from the commit order of the other tokens, we rerun the confidence-order policies with EOS allowed only at the frontier (\cref{tab:main}, right). This splits each wide-window loss into an \emph{EOS cost} (\eosf{} minus \off{}) and a \emph{content-order cost} (\fo{} minus \eosf{}). On NLD, almost all of the loss is EOS cost, on Fast-dLLM v2 almost all is content-order cost, while LLaDA has both. Keeping EOS at the frontier is therefore enough for NLD on GSM8K, but not for LLaDA, Fast-dLLM v2, or NLD-3B on HumanEval, so a fix found on one model has to be checked on the others. \Cref{app:tables} gives this split for every task.

\noindent \textbf{The reported collapse of LLaDA is mostly an EOS problem.} In CD4LM's released pipeline~\citep{liang2026cd4lm}, LLaDA-8B falls from 77.7\% on GSM8K with 32-token blocks to 13.2\% with one 256-token block. Allowing EOS only at the frontier restores 64.5\%, and committing the frontier at every step 79.3\%, with the same number of forward passes, so a decoding policy fixes most of the collapse without retraining. \Cref{app:eos} describes the pipeline and each change.

\subsection{Frontier policies recover the accuracy}
\label{sec:other}

\begin{table}[!t]
\caption{\textbf{Frontier policies recover most of the loss; only \fo{} does so on all four models.} GSM8K accuracy (\%) at the wide window (NLD $\B=128$; LLaDA $\B=256$, prompts A and B; Fast-dLLM v2 $\B=128$). Last row: \off{} at each model's default window (NLD and LLaDA $\B=32$; Fast-dLLM v2, 8-token sub-blocks).}
\label{tab:baselines}
\centering
\footnotesize
\setlength{\tabcolsep}{1.8pt}
\begin{tabular}{>{\scriptsize}lccccc}
\toprule
\textbf{Policy} & \hd{\textbf{NLD}}{\textbf{3B}} & \hd{\textbf{NLD}}{\textbf{8B}} & \hd{\textbf{LLaDA}}{\textbf{A}} & \hd{\textbf{LLaDA}}{\textbf{B}} & \hd{\textbf{Fast-}}{\textbf{dLLM v2}} \\
\midrule
\off{} & 79.1 & 87.3 & 50.6 & 35.5 & 79.4 \\
\fa{} & 85.7 & 89.6 & 67.9 & 75.4 & 82.6 \\
\gate{} & 87.3 & 90.9 & 72.6 & 80.7 & 80.8 \\
\ebud{} & 87.1 & 91.4 & 73.2 & 80.8 & 80.3 \\
\eosf{} & 86.2 & 91.0 & 54.0 & 57.4 & 79.5 \\
\fo{} & 87.6 & 90.1 & 71.6 & 82.3 & 85.4 \\
\midrule
\off{}, default & 87.6 & 92.0 & 73.2 & 80.3 & 84.8 \\
\bottomrule
\end{tabular}
\end{table}

\noindent \textbf{Which part of the policy matters.} All policies share the decoder and threshold (\cref{tab:baselines}) and differ in which ahead commits they allow: \fa{} any confident one, \gate{} any within 16 positions, and \ebud{} confident ones with little entropy to their left. On NLD and LLaDA, all three recover most of \fo{}'s gain, so the loss there comes mostly from fallback commits far ahead, often EOS. On Fast-dLLM v2, they recover about half of it or less: confident commits a few positions ahead also cause errors. A policy that works on all four models must therefore also stop confident ahead commits, as \fo{} does.

\noindent \textbf{How far ahead is safe?} A distance cap $k$ lets commits reach at most $k$ positions past the frontier ($k=0$ is \fo{}). Small caps ($k\le2$) keep \fo{}'s accuracy in most settings, and no cap value increases TPF by more than $18\%$, so there is little-to-no reason to commit ahead of the frontier at all. \Cref{app:distcap} gives the full sweep.

\noindent \textbf{Robustness.} At the wide window, \fo{} beats \off{} at every threshold from 0.7 to 0.95, under sampling, and in most code, translation and summarization settings (\cref{app:tables}). The gain grows with the window: on GSM8K, \off{} at each model's default window is within two points of \fo{} at the wide window (last row of \cref{tab:baselines}). On Fast-dLLM v2, \off{} already loses 4 points on GSM8K when its window grows from the default 8-token sub-block to the 32-token training block, and \fo{} recovers them (\cref{app:fdllm}).

\noindent \textbf{Summary (RQ2).} Wide windows lose accuracy because of the order of commits. Early EOS tokens and early content tokens are the two causes, and their mix depends on the model, so the cheapest fix does too (\eosf{} for NLD, \fa{} for LLaDA). Only \fo{} works on all four models at about the same speed, which makes it the safe default.

\section{Most failures start when the fallback commits ahead of the frontier}
\label{sec:risk}

To find which commits start the failures (RQ3), we decode each problem twice with the same model and prompt: once with \off{} and once left-to-right with \ltr{} (the \emph{reference}). An error can show up many tokens after the commit that caused it, so we look at the first commit where the two decodes differ, which we call the \emph{root}.
\begin{definitionbox}{}
\textbf{Root:} the first commit where \off{} differs from \ltr{}.
\end{definitionbox}

A \emph{harmful departure} is a problem that \ltr{} solves and \off{} fails; a \emph{helpful departure} is the reverse.

\begin{table}[!t]
\caption{\textbf{Most harmful departures start at fallback commits.} Wide windows (NLD $\B=128$, four tasks; Fast-dLLM v2 $\B=128$; LLaDA $\B=256$, prompts A and B). \emph{Fallback share}: of all commits, and of the roots of harmful departures. \emph{Rate ratios} per commit: harmful roots at fallback vs confident commits, and departures (harmful and helpful) at distance $\geq1$ vs 0 at the same confidence (range over distance bins; the reference commits the frontier, which lowers the frontier rate).}
\label{tab:risk}
\centering
\footnotesize
\setlength{\tabcolsep}{3pt}
\begin{tabular}{lcccc}
\toprule
 & \multicolumn{2}{c}{\textbf{Fallback share (\%)}} & \multicolumn{2}{c}{\textbf{Rate ratio}} \\
\cmidrule(lr){2-3}\cmidrule(lr){4-5}
\textbf{Model} & \textbf{commits} & \hd{\textbf{harmful}}{\textbf{roots}} & \hd{\textbf{fallback /}}{\textbf{confident}} & \hd{\textbf{ahead /}}{\textbf{frontier}} \\
\midrule
NLD-3B & 14 & 61 & 20.2 & 9.6--14.6 \\
NLD-8B & 35 & 72 & 31.5 & 9.0--14.3 \\
Fast-dLLM v2 & 11 & 70 & 19.1 & 7.8--12.3 \\
LLaDA & 17 & 68 & 10.3 & 1.3--1.9 \\
\bottomrule
\end{tabular}
\end{table}

\noindent \textbf{Fallback commits start most failures, mainly ahead of the frontier.} At wide windows, fallback commits are only 11--35\% of all commits, yet they start 61--72\% of the harmful departures (\cref{tab:risk}). Fallback commits also start most helpful departures, but net of these, they account for 56--69\% of the accuracy gap on three of the four models. In addition, most of these fallback roots lie ahead of the frontier. \Cref{app:risk} breaks the roots down by kind, distance and model.

\noindent \textbf{Distance adds risk at the same confidence.} Ahead commits have lower confidence than frontier commits, so their risk could come from confidence alone. To separate causes, we model the number of departures each commit starts by its distance and confidence (Poisson regression). On the three block-diffusion models, an ahead commit still starts departures several times as often as a frontier commit at the same confidence (\cref{tab:risk}). Thus, a confidence threshold alone cannot flag the risky commits, so a decoding policy should use a token's distance from the frontier as well.

\noindent \textbf{Changing one step repairs failing answers.} To test whether the root commit is the one that causes the failure, we \emph{replay} each failing decode with a single change: at the root's step, the decoder commits tokens only at the frontier. As controls, we instead change up to four earlier steps of the same answer, preferring steps that make the same kind of ahead commit as the root (fallback or confident) and come just before it. We replay NLD-3B and Fast-dLLM v2 on GSM8K and MATH-500 at $\B=128$, and NLD-3B on GSM8K at $\B=256$. In all five cases, changing the root's step repairs 7--39 points more failing answers than changing a control step (\cref{app:defer}), so the root's step contributes significantly to the failure. \fo{} makes the same change at every step, without knowing which answers fail.

\noindent \textbf{Summary (RQ3).} Failing decodes deviate from left-to-right generations mostly at fallback commits ahead of the frontier. On the block-diffusion models, distance from the frontier adds risk even at the same confidence, and changing only the root's step repairs the answer more often than changing a matched control step. This is why \fo{} works: it never lets the fallback, or an EOS token, commit ahead of the frontier.

\section{Related work}
\label{sec:related}
\noindent \textbf{Decoding policies.} Confidence and entropy are commonly used for decoding policies to decide how many tokens to commit~\citep{wu2025fastdllm,benhamu2025eb,hong2025wino,fastdllmpp2026,israel2025apd}. Analyses relate errors from independent parallel commits to the conditional total correlation of the committed positions~\citep{serialdepth2026,zhang2026infotheory}. This quantity measures dependence among the committed tokens given their context. Building on this work, we ask where to commit instead of how many and find that order has the larger effect on GSM8K accuracy.

Prior studies report nearly left-to-right generation with 32-token blocks~\citep{hou2026pard,gong2025diffucoder}. The benefit of decoding easier positions first can grow with model error~\citep{zhang2026infotheory}. Adaptive generation order also helps on Sudoku and planning tasks~\citep{kim2025worst,ye2025beyond}. PC-Sampler uses position weights, providing a softer restriction than a fixed distance cap~\citep{huang2025pcsampler}. Other policies use stability across steps to gate commits~\citep{tacg2026,fang2026frontier}. \citet{ni2026flexibility} find that arbitrary order can reduce diversity and weaken pass@$k$ scaling, while \citet{lcrtrap2026} attribute much of this loss to low-confidence remasking. \citet{zhong2026parallelism} measure parallelism and generation order in aggregate across complete decodes. Different from these works, we trace individual commits and replay selected steps to test how they affect the final answer.

Several concurrent decoding policies restrict commit positions. These include PARD~\citep{hou2026pard}, Fork-dLLM~\citep{frkovic2026fork}, RPD~\citep{he2026rpd}, and frontier gating~\citep{yeom2026answerfirst}. We compare policies based on these methods and examine why their restrictions affect speed and accuracy.

\noindent \textbf{Effect of block size.} Early termination and accuracy loss with wide-block decoding windows have been reported in several studies~\citep{liang2026cd4lm,park2026sacm,kim2026early,kim2025rainbow,liu2026voidpadding,jazbec2025unmasking,li2025daedal}. Proposed remedies change EOS probabilities, predict the answer length~\citep{survivallength2026}, or retrain the model. We instead trace the early EOS token to the fallback and fix it in the decoding policy.

\noindent \textbf{Revision and speculation.} There also exist decoding policies that allow remasking previously committed tokens~\citep{asaria2026neither,yu2026revise,ye2026sumi,nafez2026hold}. A supplementary experiment on DiffusionGemma~\citep{diffusiongemma2026} finds that tokens first committed ahead of the frontier are revised more often than tokens committed at the frontier (\cref{app:dg}). There are also drafting methods that reuse spans to accelerate generation~\citep{yang2023llma,kou2024cllms,wei2026recguide,wu2026ceiling,sahoo2026uno}. \citet{peng2025efficient} study inference efficiency under load. We find that wide-window decoding is especially expensive at large batch sizes (\cref{sec:speed}).

\section{Conclusion}
In this paper, we conduct a commit-level analysis by tracing every commit of four open-weight DLMs to find where their speed and their errors come from. The speed comes from the tokens committed at the frontier, so faster decoders should increase the number of tokens committed there in each decoding step instead of widening the window. The errors start mostly where the fallback commits ahead of the frontier, and an early EOS token is the worst case. Committing only at the frontier keeps the speed and recovers most of the accuracy, even with windows eight times the training block. This gives a simple recipe for decoding: commit only at the frontier, or at least keep the fallback and EOS there. For future models, a wider window will make decoding faster only if the model is confident further ahead of the frontier, so this is what training should aim for.

\section*{Limitations}
Four limitations bound our claims.
(i) \emph{Decoders.} We study threshold decoding that never revises a commit. Decoding policies that revise commits appear only in a side experiment on DiffusionGemma~\citep{diffusiongemma2026} (see \cref{app:dg}).
(ii) \emph{Tasks and models.} Our tasks are math, code, translation and summarization, where left-to-right order is natural. Tasks where other orders can help, such as Sudoku, planning~\citep{kim2025worst,ye2025beyond} and infilling, are out of scope. Our four main models have at most 8B parameters, and we do not study larger models.
(iii) \emph{Window size.} The accuracy effects are large mainly with windows wider than the training block (NLD is trained with $\B=32$, and our main results use $\B=128$ and 256). At the training block, \off{} loses less accuracy and \fo{} gains less (\cref{sec:other}). Our accuracy findings therefore matter most when a model is decoded with a wider window than it was trained with, as in recent evaluations of LLaDA with one 256-token block~\citep{liang2026cd4lm}. Fast-dLLM v2 is an exception, as it loses accuracy as its window widens even inside its training block (\cref{app:fdllm}).
(iv) \emph{Roots and replays.} We pick roots after we know the outcome, and the control steps reduce this selection effect but do not remove it. Our Poisson model does not account for how hard each problem is, and we ran replays only on NLD-3B and Fast-dLLM v2. The roots therefore show where failures start, but the evidence that the root's step causes the failure is limited to these two models.

\bibliography{refs}

\appendix
\crefalias{section}{appendix}
\crefalias{subsection}{appendix}
\makeatletter
\setlength{\@fptop}{0pt}\setlength{\@fpsep}{12pt}\setlength{\@fpbot}{0pt plus 1fil}
\setlength{\@dblfptop}{0pt}\setlength{\@dblfpsep}{12pt}\setlength{\@dblfpbot}{0pt plus 1fil}
\makeatother

\section{Setup details}
\label{app:setup}

\Cref{tab:datasets} lists the tasks.

\begin{table*}[tp]
\caption{\textbf{Tasks.} Size: number of problems. Answer length: mean over the \off{} answers of NLD-3B at $\B=32$ (tokens for math and code, words for translation and summarization). The CD4LM experiment (\cref{sec:eos}) uses that work's own 440-problem GSM8K subset.}
\label{tab:datasets}
\centering
\footnotesize
\begin{tabularx}{\textwidth}{@{}>{\raggedright\arraybackslash}p{3.1cm}>{\raggedright\arraybackslash}Xr>{\raggedright\arraybackslash}p{1.9cm}>{\raggedright\arraybackslash}p{2.3cm}l@{}}
\toprule
\textbf{Task} & \textbf{Source and split} & \textbf{Size} & \textbf{Metric} & \textbf{Models} & \textbf{Answer length} \\
\midrule
GSM8K~\citep{cobbe2021gsm8k} & openai/gsm8k, test & 1319 & accuracy & all four & 179 tokens \\
MATH-500~\citep{hendrycks2021math,lightman2023verify} & HuggingFaceH4/MATH-500 & 500 & accuracy & NLD, Fast-dLLM v2 & 591 tokens \\
HumanEval~\citep{chen2021codex} & EvalPlus~\citep{liu2023evalplus}, base tests & 164 & pass@1 & all four & 552 tokens \\
MBPP~\citep{austin2021mbpp} & EvalPlus~\citep{liu2023evalplus}, base tests & 378 & pass@1 & all four & 461 tokens \\
WMT16 de$\to$en~\citep{bojar2016wmt} & wmt/wmt16, test, 500 random sentences & 500 & chrF~\citep{popovic2015chrf} & NLD-3B, LLaDA & 18 words \\
CNN/DailyMail~\citep{hermann2015cnndm,see2017pointer} & abisee/cnn\_dailymail 3.0.0, test, 500 random articles & 500 & ROUGE-L~\citep{lin2004rouge} & NLD-3B, LLaDA & 72 words \\
\bottomrule
\end{tabularx}
\end{table*}

\noindent \textbf{Baseline policies.} \gate{} applies \off{} only to the masked positions $j<f+16$, where $f$ is the frontier~\citep[default width]{yeom2026answerfirst}. \ebud{} keeps RPD's entropy budget ($\beta=4$ nats, summed over the masked positions to the left that are not committed in the same step) and its fallback (the most confident of the first 32 masked positions). It leaves out RPD's layer-stability test, which adds candidates with confidence between 0.6 and 0.9~\citep{he2026rpd}. \Cref{app:tables} compares it with RPD's released decoder.

\noindent \textbf{Grading.} We grade math answers with math-verify and code with the EvalPlus base tests~\citep{liu2023evalplus}. NLD-8B and Fast-dLLM v2 often add an example code block after the function, so we grade the block that defines the function instead of the last block. This matters only on NLD-8B MBPP, where grading the last block would put \fo{} and \ltr{} about 5 points below \off{} instead of about 2 points above it. With the stricter EvalPlus tests, every difference larger than 1.1 points keeps its sign.

\noindent \textbf{Hardware.} All runs used NVIDIA A100-40GB GPUs.

\section{Confidence and speed}
\label{app:conf}

\subsection{Confidence by distance and tokens per step}
\label{app:prop}

\noindent \textbf{First confidence by distance.} \Cref{fig:speed3}a uses the left-to-right decodes of \cref{tab:main} (\ltr{}; on Fast-dLLM v2, sub-block 1), where the forward pass predicts a token and a confidence for every masked position but the decoder commits only the frontier. We leave out the positions that \off{}'s fallback would have picked before they ever passed the threshold (0--25\%), and average the shares per problem. Across models, tasks and windows, 52--76\% of positions first pass the threshold within seven positions of the frontier. In these GSM8K runs, 60--99\% of the positions that become confident 8--31 positions ahead, by model and distance, predict the token that the model writes there later, and 25--84\% of those further ahead.

\begin{figure}[!htb]
 \centering
 \includegraphics[width=\columnwidth,trim={0 0 222bp 11bp},clip]{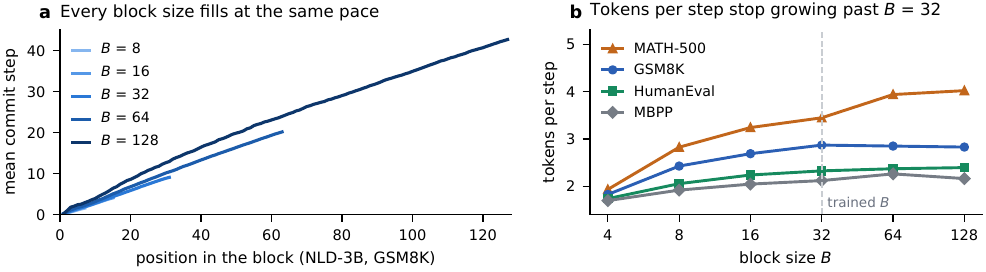}
 \vspace{-1.6em}
 \caption{\textbf{A larger block mainly adds positions.} Mean commit step by position for $\B=8$--128 (NLD-3B, GSM8K, \off{}); the curves nearly coincide.}
 \label{fig:front}
\end{figure}

\noindent \textbf{Tokens per step by block size.} On NLD-3B, a position at a given offset in its block is committed at about the same step for every $\B$ from 8 to 128: the mean commit step grows by 0.24--0.34 steps per position on GSM8K and 0.19--0.24 on MATH-500 (\cref{fig:front}). A larger block therefore mainly adds positions, and tokens per step (commits divided by steps) stop growing. From $\B=32$ to 128, they change by a factor of 0.99--1.03 on NLD-3B (MATH-500: 1.17; \cref{fig:speed3}b), while NLD-8B still gains 10--17\% on GSM8K and MATH-500.

\noindent \textbf{Other models.} LLaDA commits 2.7 tokens per step at $\B=8$ and 3.0--3.3 at $\B=32$ to 128 (GSM8K, prompt A; prompt B: 2.4--2.7), and Fast-dLLM v2's TPF stops growing at $\B=64$ (\cref{tab:grid}). \fo{} behaves the same way: from $\B=32$ to the wide windows, its TPF changes by a factor of 0.96--1.24, and by more than 1.10 only on NLD-8B.

\subsection{Speed and compute}
\label{app:speed}

{\small\begin{table*}[p]
\centering
\scriptsize
\caption{\textbf{All model $\times$ task $\times$ window settings (greedy).} Ahead: ahead commits among the commits of \off{} (\%). Accuracy (\%) of \off{}, \seq{} and \fo{}; TPF: tokens per forward. Window: $\B$ for NLD and LLaDA (256 = one block); for Fast-dLLM v2, the sub-block for 4--16, the whole 32-token training block at 32, and $\B$ for 64 and 128. A dash marks a setting we did not run.}\label{tab:grid}
\begin{tabular}{lllrrrrrr}
\toprule
Model & Task & Window & Ahead & \off{} & \seq{} & \fo{} & TPF \off{} & TPF \fo{} \\
\midrule

NLD-3B & GSM8K & 4 & 6 & 88.6 & 88.9 & -- & 1.51 & -- \\
NLD-3B & GSM8K & 8 & 13 & 89.1 & 89.2 & -- & 2.06 & -- \\
NLD-3B & GSM8K & 16 & 23 & 87.9 & 88.8 & -- & 2.46 & -- \\
NLD-3B & GSM8K & 32 & 32 & 87.6 & 88.3 & 88.1 & 2.75 & 2.61 \\
NLD-3B & GSM8K & 64 & 41 & 87.0 & 88.9 & 88.4 & 2.72 & 2.74 \\
NLD-3B & GSM8K & 128 & 49 & 79.1 & 79.7 & 87.6 & 2.27 & 2.77 \\
NLD-3B & GSM8K & 256 & 52 & 54.5 & -- & 88.1 & 1.24 & 2.51 \\
NLD-3B & MATH-500 & 4 & 6 & 77.0 & 75.4 & -- & 1.56 & -- \\
NLD-3B & MATH-500 & 8 & 11 & 72.8 & 76.0 & -- & 2.30 & -- \\
NLD-3B & MATH-500 & 16 & 18 & 70.0 & 75.2 & -- & 2.84 & -- \\
NLD-3B & MATH-500 & 32 & 28 & 74.4 & 75.6 & 75.2 & 3.18 & 3.18 \\
NLD-3B & MATH-500 & 64 & 35 & 71.0 & 75.6 & 74.8 & 3.60 & 3.35 \\
NLD-3B & MATH-500 & 128 & 43 & 69.6 & 73.8 & 73.2 & 3.54 & 3.38 \\
NLD-3B & MATH-500 & 256 & 48 & 64.0 & -- & 72.8 & 2.81 & 3.31 \\
NLD-3B & HumanEval & 4 & 7 & 76.8 & 76.2 & -- & 1.46 & -- \\
NLD-3B & HumanEval & 8 & 15 & 72.0 & 72.6 & -- & 1.81 & -- \\
NLD-3B & HumanEval & 16 & 24 & 78.0 & 76.8 & -- & 2.04 & -- \\
NLD-3B & HumanEval & 32 & 31 & 73.2 & 75.0 & 79.3 & 2.12 & 2.12 \\
NLD-3B & HumanEval & 64 & 40 & 70.7 & 75.6 & 78.7 & 2.12 & 2.19 \\
NLD-3B & HumanEval & 128 & 45 & 64.6 & 70.1 & 81.1 & 1.98 & 2.21 \\
NLD-3B & MBPP & 4 & 7 & 72.5 & 72.8 & -- & 1.44 & -- \\
NLD-3B & MBPP & 8 & 16 & 72.5 & 72.5 & -- & 1.71 & -- \\
NLD-3B & MBPP & 16 & 25 & 72.2 & 74.6 & -- & 1.89 & -- \\
NLD-3B & MBPP & 32 & 34 & 71.2 & 74.6 & 74.9 & 1.95 & 2.05 \\
NLD-3B & MBPP & 64 & 40 & 66.4 & 70.4 & 71.2 & 2.03 & 2.22 \\
NLD-3B & MBPP & 128 & 49 & 63.5 & 64.3 & 72.5 & 1.82 & 2.20 \\
NLD-8B & GSM8K & 4 & 5 & 92.0 & -- & -- & 1.57 & -- \\
NLD-8B & GSM8K & 8 & 10 & 92.2 & -- & -- & 2.25 & -- \\
NLD-8B & GSM8K & 16 & 16 & 92.5 & -- & -- & 2.86 & -- \\
NLD-8B & GSM8K & 32 & 26 & 92.0 & 92.0 & 91.3 & 3.17 & 3.26 \\
NLD-8B & GSM8K & 64 & 35 & 91.1 & -- & -- & 3.17 & -- \\
NLD-8B & GSM8K & 128 & 37 & 87.3 & 86.7 & 90.1 & 3.37 & 3.78 \\
NLD-8B & GSM8K & 256 & 44 & 73.8 & -- & 88.8 & 3.50 & 4.05 \\
NLD-8B & MATH-500 & 4 & 5 & 87.0 & -- & -- & 1.57 & -- \\
NLD-8B & MATH-500 & 8 & 10 & 85.2 & -- & -- & 2.25 & -- \\
NLD-8B & MATH-500 & 16 & 17 & 87.0 & -- & -- & 2.73 & -- \\
NLD-8B & MATH-500 & 32 & 25 & 84.0 & 85.8 & 85.6 & 3.24 & 3.20 \\
NLD-8B & MATH-500 & 64 & 34 & 85.4 & -- & -- & 3.37 & -- \\
NLD-8B & MATH-500 & 128 & 41 & 81.2 & 82.4 & 85.0 & 3.63 & 3.40 \\
NLD-8B & MATH-500 & 256 & 46 & 79.4 & -- & 82.4 & 3.61 & 3.63 \\
NLD-8B & HumanEval & 4 & 6 & 86.0 & -- & -- & 1.54 & -- \\
NLD-8B & HumanEval & 8 & 12 & 84.1 & -- & -- & 2.08 & -- \\
NLD-8B & HumanEval & 16 & 22 & 83.5 & -- & -- & 2.22 & -- \\
NLD-8B & HumanEval & 32 & 28 & 83.5 & 82.3 & 83.5 & 2.74 & 2.52 \\
NLD-8B & HumanEval & 64 & 38 & 82.9 & -- & -- & 2.68 & -- \\
NLD-8B & HumanEval & 128 & 53 & 86.0 & 86.0 & 84.8 & 2.15 & 2.64 \\
NLD-8B & MBPP & 4 & 8 & 85.7 & -- & -- & 1.45 & -- \\
NLD-8B & MBPP & 8 & 19 & 84.9 & -- & -- & 1.51 & -- \\
NLD-8B & MBPP & 16 & 28 & 84.7 & -- & -- & 1.64 & -- \\
NLD-8B & MBPP & 32 & 29 & 83.1 & 83.9 & 82.8 & 2.15 & 2.04 \\
NLD-8B & MBPP & 64 & 37 & 82.3 & -- & -- & 2.12 & -- \\
NLD-8B & MBPP & 128 & 38 & 82.3 & 81.7 & 83.9 & 2.01 & 2.33 \\
LLaDA (A) & GSM8K & 32 & 35 & 73.2 & -- & -- & 3.32 & -- \\
LLaDA (A) & GSM8K & 256 & 74 & 50.6 & -- & 71.6 & 2.95 & 3.13 \\
LLaDA (B) & GSM8K & 32 & 36 & 80.3 & -- & -- & 2.85 & -- \\
LLaDA (B) & GSM8K & 256 & 72 & 35.5 & -- & 82.3 & 1.76 & 2.72 \\
Fast-dLLM v2 & GSM8K & 4 & 6 & 86.4 & -- & -- & 2.28 & -- \\
Fast-dLLM v2 & GSM8K & 8 & 12 & 84.8 & -- & 86.7 & 2.85 & 2.78 \\
Fast-dLLM v2 & GSM8K & 16 & 20 & 81.8 & -- & -- & 3.19 & -- \\
Fast-dLLM v2 & GSM8K & 32 & 30 & 80.8 & -- & 86.1 & 3.38 & 3.20 \\
Fast-dLLM v2 & GSM8K & 64 & 40 & 79.2 & -- & 85.7 & 3.56 & 3.34 \\
Fast-dLLM v2 & GSM8K & 128 & 50 & 79.4 & 82.0 & 85.4 & 3.32 & 3.14 \\
Fast-dLLM v2 & MATH-500 & 4 & 6 & 61.4 & -- & -- & 2.28 & -- \\
Fast-dLLM v2 & MATH-500 & 8 & 11 & 60.4 & -- & 66.4 & 2.92 & 2.78 \\
Fast-dLLM v2 & MATH-500 & 16 & 19 & 54.6 & -- & -- & 3.15 & -- \\
Fast-dLLM v2 & MATH-500 & 32 & 27 & 52.6 & -- & 62.0 & 3.34 & 3.11 \\
Fast-dLLM v2 & MATH-500 & 64 & 37 & 52.2 & -- & 62.2 & 3.52 & 3.32 \\
Fast-dLLM v2 & MATH-500 & 128 & 46 & 52.2 & -- & 62.0 & 3.41 & 3.27 \\
\bottomrule
\end{tabular}
\end{table*}
}

\noindent \textbf{Ahead commits and TPF.} \Cref{tab:grid} gives the ahead share and the TPF of every setting. Both count only answer tokens, including the EOS token that ends the answer, and leave out the EOS padding after it, which \off{} also commits: on NLD-3B GSM8K, padding is 9\%, 32\% and 55\% of its commits at $\B=32$, 128 and 256. Over the 34 settings of \cref{fig:speed3}c, \fo{} reaches 0.92--2.03 times the TPF of \off{}; with Fast-dLLM v2's code runs added (40 settings), 0.88--2.03, with the lowest value on MBPP at $\B=32$. The largest ratios come from answers that \off{} cuts short: on problems whose two answers differ in length by at most 10\%, the range is 0.88--1.27.

\noindent \textbf{Compute.} Each NLD forward pass processes all $\B$ positions of the block, so it processes $\B/\mathrm{TPF}$ positions per answer token (\cref{tab:grid}). On NLD-3B GSM8K, this is 12 at $\B=32$ for both policies, 46 for \fo{} and 57 for \off{} at 128, and 102 and 207 at 256. Across NLD settings, it grows 3.4--3.8 times for \fo{} from $\B=32$ to 128, and 6.4--8.3 times to 256. LLaDA processes the prompt and the whole 256-token answer in every forward pass: 112--149 positions per answer token for \fo{} and 118--231 for \off{}.

\noindent \textbf{Wall-clock time.} \Cref{tab:steady} gives NLD-3B throughput. At batch size 1, the GPU has spare capacity, so the extra positions of a wider window cost almost nothing: $\B=128$ is even 6--11\% faster than $\B=32$. At batch size 32, the GPU is fully used, and every extra position costs time. On Fast-dLLM v2 at batch size 1, \fo{} takes 0.77--1.12 times as long as \off{} per answer token (two runs shared each GPU).

\begin{table}[!htb]
\caption{\textbf{NLD-3B throughput relative to AR decoding on the same GPU.} Steady-state throughput over the first 512 tokens of 64 long MATH-500 answers (A100-40GB, eager PyTorch). The self-speculation and $\B=32$ rows come from one benchmark. The larger-block rows come from a second benchmark of two jobs run in opposite order, which agree within 3\%.}
\label{tab:steady}
\centering
\footnotesize
\setlength{\tabcolsep}{4pt}
\begin{tabular}{lccc}
\toprule
Decoder & batch 1 & batch 8 & batch 32 \\
\midrule
self-speculation & 4.72 & 3.42 & 1.93 \\
\quad + LoRA drafter & 4.83 & 3.60 & 2.00 \\
\multicolumn{4}{@{}l}{\emph{threshold decoding}} \\
\off{}, $\B=32$ & 2.95 & 1.57 & 0.63 \\
\fo{}, $\B=32$ & 2.85 & 1.54 & 0.62 \\
\off{}, $\B=128$ & 3.12 & 0.88 & 0.25 \\
\fo{}, $\B=128$ & 3.16 & 0.91 & 0.26 \\
\off{}, $\B=256$ & 2.76 & 0.47 & 0.14 \\
\fo{}, $\B=256$ & 2.87 & 0.46 & 0.13 \\
\bottomrule
\end{tabular}
\end{table}

\section{Accuracy}
\label{app:acc}

\subsection{Every task, threshold and policy}
\label{app:tables}

\noindent \textbf{Order and parallelism on every task.} \Cref{tab:factorial_full} repeats the comparison of \cref{tab:main}, with intervals, in every setting where we ran all four policies. \Cref{tab:eosctl} repeats it with EOS allowed only at the frontier in every cell, which needed new runs of \seq{} only.

\noindent \textbf{Frontier policies.} \Cref{tab:nld3b_rules} compares the policies on the four NLD-3B tasks. \fa{} recovers 78--82\% of \fo{}'s GSM8K gain on NLD, but on NLD-3B only 52\% on HumanEval, 20\% on MBPP and none on MATH-500, and on HumanEval \gate{} and \ebud{} also stay 10 points below \fo{}. RPD's released decoder, which keeps the layer-stability test that our \ebud{} leaves out, matches \fo{} on LLaDA (73.2 and 81.1 against 71.6 and 82.3 for prompts A and B) at 15\% more TPF.

\noindent \textbf{Code and the training block.} On code at the wide window, \fo{} gains 9--17 points over \off{} on NLD-3B and Fast-dLLM v2 and 1--3 points on LLaDA, and stays within two points on NLD-8B. At the training block ($\B=32$), it changes accuracy by at most 1.6 points on NLD math and NLD-8B code, but still gains 4--6 points on NLD-3B code.

\noindent \textbf{Thresholds.} \Cref{tab:thr} varies the threshold. At every threshold from 0.7 to 0.95, the gap between \fo{} and \off{} at the wide window is significant (the lowest bound of its 95\% interval is 0.2 points, on MATH-500); at $\B=32$, it is significant only at 0.7 and 0.8. A lower threshold commits more tokens per step, so we also compare the two policies at the same speed. On NLD-3B GSM8K at threshold 0.8, the two policies have about the same TPF and \fo{} is 9.6 points more accurate, but on MATH-500 the gap can vanish: \off{} at 0.8 and \fo{} at 0.7 both reach 4.4 TPF and 67\%.

\noindent \textbf{Sampling.} With three seeds at temperature 0.7, \fo{} beats \off{} on GSM8K at the wide windows by $7.2\pm1.6$ points on NLD-3B ($\B=128$), 19.5 on NLD-8B ($\B=256$), 44.5 on LLaDA and 5.3 on Fast-dLLM v2.

\noindent \textbf{Translation and summarization.} On NLD-3B at $\B=128$, \fo{} beats \off{} by $5.7\pm1.3$ chrF and $7.5\pm1.0$ ROUGE-L (paired differences of sentence-level scores), and \seq{} changes either score by at most 0.2. Under \off{}, 83\% of the summaries are less than half as long as at the narrow window; under \fo{}, none are. On LLaDA at $\B=256$, \fo{} raises ROUGE-L from 2.4 to 24.6, because every \off{} summary ends almost at once, but changes translation little ($+0.8\pm1.1$ chrF): the reference translations are short (median 17 words), and only 2\% of \off{} translations are cut short, so there is little early EOS to fix.

\begin{table*}[tp]
\caption{\textbf{NLD-3B accuracy (\%) by task and policy.} Policies at $\B=128$, and \off{} at $\B=32$ for comparison; \cref{tab:factorial_full} gives \seq{} and \ltr{}.}
\label{tab:nld3b_rules}
\centering
\small
\setlength{\tabcolsep}{3pt}
\begin{tabular}{lccccccc}
\toprule
Task & {\scriptsize\off{}} & {\scriptsize\fo{}} & {\scriptsize\fa{}} & {\scriptsize\gate{}} & {\scriptsize\ebud{}} & {\scriptsize\eosf{}} & {\scriptsize\off{}}, $\B=32$ \\
\midrule
GSM8K & 79.1 & 87.6 & 85.7 & 87.3 & 87.1 & 86.2 & 87.6 \\
MATH-500 & 69.6 & 73.2 & 68.8 & 71.8 & 72.2 & 70.8 & 74.4 \\
HumanEval & 64.6 & 81.1 & 73.2 & 70.7 & 71.3 & 70.1 & 73.2 \\
MBPP & 63.5 & 72.5 & 65.3 & 72.0 & 71.7 & 69.8 & 71.2 \\
\bottomrule
\end{tabular}
\end{table*}

\begin{table*}[tp]
\caption{\textbf{Order $\times$ parallelism in every setting with all four policies (accuracy, \%; effects in points, $\pm$ half the width of the 95\% problem-bootstrap interval; \cref{sec:factorial}).} \off{}: confidence order, several tokens per step; \seq{}: confidence order, one; \ltr{}: frontier order, one; \fo{}: frontier order, several.}
\label{tab:factorial_full}
\centering
\scriptsize
\setlength{\tabcolsep}{1.7pt}
\begin{tabular}{llccccccc}
\toprule
Model, window & task & \off{} & \seq{} & \ltr{} & \fo{} & order & parallelism & interaction \\
\midrule
NLD-3B, $\B=128$ & GSM8K & 79.1 & 79.7 & 88.0 & 87.6 & $+$8.4 $\pm$ 1.7 & $-$0.5 $\pm$ 0.8 & $+$0.2 $\pm$ 1.5 \\
NLD-3B, $\B=128$ & MATH-500 & 69.6 & 73.8 & 76.6 & 73.2 & $+$3.2 $\pm$ 2.9 & $-$3.8 $\pm$ 2.0 & $+$0.8 $\pm$ 3.3 \\
NLD-3B, $\B=128$ & HumanEval & 64.6 & 70.1 & 80.5 & 81.1 & $+$13.4 $\pm$ 6.3 & $-$2.4 $\pm$ 2.7 & $+$6.1 $\pm$ 5.2 \\
NLD-3B, $\B=128$ & MBPP & 63.5 & 64.3 & 73.8 & 72.5 & $+$9.3 $\pm$ 4.4 & $-$1.1 $\pm$ 1.9 & $-$0.5 $\pm$ 3.6 \\
NLD-3B, $\B=256$ & GSM8K & 54.5 & 55.0 & 88.0 & 88.1 & $+$33.3 $\pm$ 2.6 & $-$0.2 $\pm$ 1.0 & $+$0.6 $\pm$ 2.0 \\
NLD-8B, $\B=128$ & GSM8K & 87.3 & 86.7 & 90.2 & 90.1 & $+$3.2 $\pm$ 1.5 & $+$0.2 $\pm$ 0.9 & $-$0.6 $\pm$ 1.8 \\
NLD-8B, $\B=128$ & MATH-500 & 81.2 & 82.4 & 83.8 & 85.0 & $+$2.6 $\pm$ 2.6 & $+$0.0 $\pm$ 2.1 & $+$2.4 $\pm$ 3.9 \\
NLD-8B, $\B=128$ & HumanEval & 86.0 & 86.0 & 85.4 & 84.8 & $-$0.9 $\pm$ 4.0 & $-$0.3 $\pm$ 1.8 & $-$0.6 $\pm$ 3.7 \\
NLD-8B, $\B=128$ & MBPP & 82.3 & 81.7 & 84.1 & 83.9 & $+$2.0 $\pm$ 2.6 & $+$0.1 $\pm$ 0.7 & $-$0.8 $\pm$ 1.7 \\
NLD-8B, $\B=256$ & GSM8K & 73.8 & 73.3 & 88.9 & 88.8 & $+$15.3 $\pm$ 2.3 & $+$0.2 $\pm$ 1.0 & $-$0.6 $\pm$ 2.1 \\
LLaDA, A, $\B=256$ & GSM8K & 50.6 & 49.9 & 71.9 & 71.6 & $+$21.5 $\pm$ 2.8 & $+$0.2 $\pm$ 0.9 & $-$1.1 $\pm$ 1.9 \\
LLaDA, B, $\B=256$ & GSM8K & 35.5 & 36.2 & 82.2 & 82.3 & $+$46.4 $\pm$ 2.9 & $-$0.3 $\pm$ 0.8 & $+$0.9 $\pm$ 1.7 \\
Fast-dLLM v2, $\B=128$ & GSM8K & 79.4 & 82.0 & 86.6 & 85.4 & $+$5.3 $\pm$ 1.9 & $-$1.9 $\pm$ 0.9 & $+$1.4 $\pm$ 1.7 \\
Fast-dLLM v2, $\B=128$ & MATH-500 & 52.2 & 59.6 & 66.0 & 62.0 & $+$8.1 $\pm$ 3.1 & $-$5.7 $\pm$ 2.2 & $+$3.4 $\pm$ 4.2 \\
\bottomrule
\end{tabular}
\end{table*}

\begin{table*}[tp]
\caption{\textbf{Order $\times$ parallelism with EOS allowed only at the frontier in every cell:} \off{} becomes \eosf{}, and \seq{} is rerun with the same restriction; \ltr{} and \fo{} commit only at the frontier, where the restriction changes nothing. Accuracy (\%) and effects (points, 95\% problem bootstrap). Par.\ and int.: parallelism and interaction effects. Kept: the order effect with the restriction as a share of the order effect without it; it is negative where the effect changes sign.}
\label{tab:eosctl}
\centering
\scriptsize
\setlength{\tabcolsep}{2.5pt}
\resizebox{\ifdim\width>\textwidth\textwidth\else\width\fi}{!}{\begin{tabular}{lllccccccccc}
\toprule
 & & & \multicolumn{4}{c}{accuracy, EOS only at the frontier} & \multicolumn{2}{c}{order effect} & & & \\
\cmidrule(lr){4-7}\cmidrule(lr){8-9}
Model & window & task & \off{} & \seq{} & \ltr{} & \fo{} & without & with & par. & int. & kept \\
\midrule
NLD-3B & $\B=$128 & GSM8K & 86.2 & 87.3 & 88.0 & 87.6 & +8.4 $\pm$ 1.7 & +1.1 $\pm$ 1.3 & $-$0.8 $\pm$ 0.6 & +0.6 $\pm$ 1.1 & 13\% \\
NLD-3B & $\B=$128 & MATH-500 & 70.8 & 74.2 & 76.6 & 73.2 & +3.2 $\pm$ 2.9 & +2.4 $\pm$ 2.8 & $-$3.4 $\pm$ 1.8 & +0.0 $\pm$ 2.7 & 75\% \\
NLD-3B & $\B=$128 & HumanEval & 70.1 & 73.2 & 80.5 & 81.1 & +13.4 $\pm$ 6.3 & +9.1 $\pm$ 5.3 & $-$1.2 $\pm$ 2.4 & +3.7 $\pm$ 5.2 & 68\% \\
NLD-3B & $\B=$128 & MBPP & 69.8 & 72.2 & 73.8 & 72.5 & +9.3 $\pm$ 4.4 & +2.1 $\pm$ 4.0 & $-$1.9 $\pm$ 1.7 & +1.1 $\pm$ 3.0 & 23\% \\
NLD-3B & $\B=$256 & GSM8K & 87.2 & 88.0 & 88.0 & 88.1 & +33.3 $\pm$ 2.6 & +0.5 $\pm$ 1.3 & $-$0.4 $\pm$ 0.6 & +0.9 $\pm$ 1.1 & 1\% \\
NLD-8B & $\B=$128 & GSM8K & 91.0 & 91.0 & 90.2 & 90.1 & +3.2 $\pm$ 1.5 & $-$0.8 $\pm$ 1.2 & $-$0.0 $\pm$ 0.8 & $-$0.1 $\pm$ 1.5 & $-$25\% \\
NLD-8B & $\B=$128 & MATH-500 & 82.6 & 84.4 & 83.8 & 85.0 & +2.6 $\pm$ 2.6 & +0.9 $\pm$ 2.5 & $-$0.3 $\pm$ 1.7 & +3.0 $\pm$ 3.5 & 35\% \\
NLD-8B & $\B=$256 & GSM8K & 90.4 & 90.2 & 88.9 & 88.8 & +15.3 $\pm$ 2.3 & $-$1.4 $\pm$ 1.3 & +0.0 $\pm$ 0.8 & $-$0.3 $\pm$ 1.6 & $-$9\% \\
LLaDA, prompt A & $\B=$256 & GSM8K & 54.0 & 53.9 & 71.9 & 71.6 & +21.5 $\pm$ 2.8 & +17.8 $\pm$ 2.7 & $-$0.1 $\pm$ 0.8 & $-$0.4 $\pm$ 1.6 & 83\% \\
LLaDA, prompt B & $\B=$256 & GSM8K & 57.4 & 57.5 & 82.2 & 82.3 & +46.4 $\pm$ 2.9 & +24.8 $\pm$ 2.6 & +0.0 $\pm$ 0.8 & +0.3 $\pm$ 1.7 & 53\% \\
Fast-dLLM v2 & $\B=$128 & GSM8K & 79.5 & 82.0 & 86.6 & 85.4 & +5.3 $\pm$ 1.9 & +5.3 $\pm$ 1.9 & $-$1.8 $\pm$ 0.9 & +1.3 $\pm$ 1.7 & 99\% \\
Fast-dLLM v2 & $\B=$128 & MATH-500 & 52.6 & 58.8 & 66.0 & 62.0 & +8.1 $\pm$ 3.1 & +8.3 $\pm$ 3.2 & $-$5.1 $\pm$ 2.2 & +2.2 $\pm$ 4.2 & 102\% \\
\bottomrule
\end{tabular}}
\end{table*}

\begin{table*}[tp]
\caption{\textbf{Threshold sweep.} Accuracy (\%) of \off{} / \fo{} with the gap (\fo{} minus \off{}, points), and below it the TPF of \off{} / \fo{}. NLD-3B; LLaDA $\B=256$ (prompt B); Fast-dLLM v2 $\B=128$; GSM8K unless stated.}
\label{tab:thr}
\centering
\small
\setlength{\tabcolsep}{4pt}
\begin{tabular}{lcccc}
\toprule
Setting & 0.7 & 0.8 & 0.9 & 0.95 \\
\midrule
NLD-3B, $\B=128$ & 73.9 / 85.5 (+11.6) & 77.3 / 86.9 (+9.6) & 79.1 / 87.6 (+8.5) & 79.5 / 87.9 (+8.4) \\
\quad TPF & 4.0 / 3.8 & 3.0 / 3.2 & 2.3 / 2.8 & 1.8 / 2.5 \\
NLD-3B, $\B=128$, MATH-500 & 61.0 / 67.2 (+6.2) & 67.0 / 70.8 (+3.8) & 69.6 / 73.2 (+3.6) & 70.2 / 75.0 (+4.8) \\
\quad TPF & 5.2 / 4.4 & 4.4 / 4.1 & 3.5 / 3.4 & 3.1 / 3.0 \\
NLD-3B, $\B=32$ & 81.0 / 85.4 (+4.4) & 84.7 / 87.1 (+2.4) & 87.6 / 88.1 (+0.5) & 88.6 / 88.2 ($-0.4$) \\
\quad TPF & 4.0 / 3.4 & 3.3 / 3.1 & 2.7 / 2.6 & 2.4 / 2.3 \\
LLaDA, $\B=256$ & 39.2 / 80.7 (+41.5) & 35.4 / 81.8 (+46.4) & 35.5 / 82.3 (+46.9) & 35.9 / 82.3 (+46.3) \\
\quad TPF & 4.2 / 4.0 & 2.7 / 3.3 & 1.8 / 2.7 & 1.3 / 2.3 \\
Fast-dLLM v2, $\B=128$ & 68.6 / 81.2 (+12.6) & 75.1 / 85.0 (+9.9) & 79.4 / 85.4 (+6.1) & 79.8 / 86.5 (+6.7) \\
\quad TPF & 5.1 / 4.4 & 4.1 / 3.8 & 3.3 / 3.1 & 2.8 / 2.7 \\
\bottomrule
\end{tabular}
\end{table*}

\subsection{Early EOS}
\label{app:eos}

\begin{figure*}[tp]
 \centering
 \includegraphics[width=\textwidth]{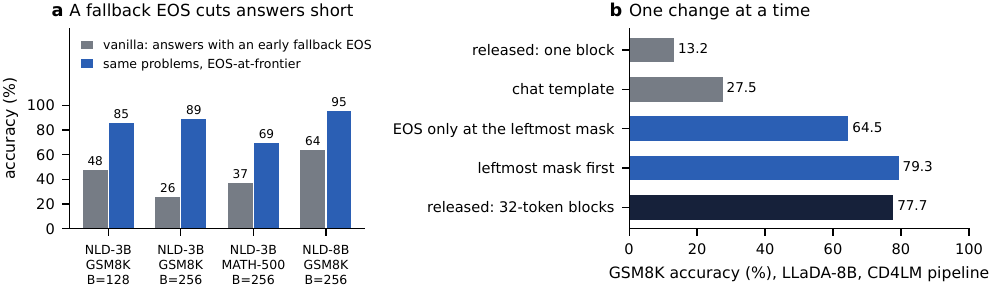}
 \vspace{-1.6em}
 \caption{\textbf{Early EOS commits cause most of the wide-window loss on NLD and most of LLaDA's reported collapse.} \textbf{(a)} NLD: accuracy of the answers whose EOS token the fallback committed early, and of the same problems under \eosf{}. \textbf{(b)} LLaDA-8B, GSM8K, in the released CD4LM evaluation pipeline (decoder, prompt and grader; 440 problems). Each of the three middle bars changes one setting of the released one-block setup; committing the leftmost masked position first removes the drop.}
\label{fig:eos}
\end{figure*}

\noindent \textbf{Early EOS on NLD-3B.} An EOS token is early when it is committed while a position to its left is still masked. In NLD's decoder, a block is finished once no position before its first EOS is masked, and a block that holds an EOS ends the answer, so an early EOS fixes the remaining length before the reasoning is written. On GSM8K at $\B=256$, 80\% of answers commit their EOS early, about half of them through the fallback. Answers with a fallback early EOS are 26\% correct on GSM8K at $\B=256$, 48\% at $\B=128$ and 37\% on MATH-500 at $\B=256$, against 89\%, 85\% and 69\% of the same problems under \eosf{} (\cref{fig:eos}a), so a fallback early EOS costs 32--63 points. A confident early EOS costs 3--14 points.

\noindent \textbf{Where the fallback puts EOS.} Fallback EOS commits have confidence of about 0.64--0.70 and lie a median 22--31 positions ahead of the frontier, which fits the padding explanation of \cref{sec:eos}: in training, positions far beyond the written text are often padding, so EOS is the model's best guess there, though not a confident one. The worst case is an EOS 1--4 positions after the frontier, which the fallback commits in 73--89\% of cases: on GSM8K at $\B=256$, 139 answers have one, and 1\% of them are correct, against 84\% under \eosf{}.

\noindent \textbf{Smaller windows.} At $\B\leq64$, 18--75\% of answers still commit their EOS ahead of the frontier, a median 5--20 positions ahead, but their lengths and accuracy match \eosf{} (median length ratio 0.97--1.00; accuracy within one point on GSM8K and MATH-500 at $\B=32$). In a small window, EOS can only be placed a few positions ahead, close to where the answer ends anyway.

\noindent \textbf{LLaDA in the CD4LM pipeline.} CD4LM reports LLaDA-8B at 13.8\% on GSM8K with one 256-token block and 77.4\% with 32-token blocks~\citep{liang2026cd4lm}, and LLaDA's authors trace such losses to the EOS padding in its fine-tuning data~\citep[Sec.~3.2]{nie2025llada}. In the released evaluation pipeline (440 problems), whose decoder commits the single most confident position per step, we get 13.2\% and 77.7\%. \Cref{fig:eos}b changes one setting at a time: allowing EOS only at the frontier gives 64.5\%, committing the frontier at every step 79.3\% (a gain of 66.1 points, $p=3\times10^{-83}$, with the same number of forward passes), and changing only the chat template 27.5\%. Most of the collapse therefore comes from where EOS is placed, and about 15 more points from the order of the content tokens.

\noindent \textbf{Several leftmost positions per step.} With two or four tokens per step, committing the leftmost positions still beats committing the most confident ones on GSM8K by 36--60 points, but without a confidence check it costs accuracy on code: on MBPP, the two leftmost positions per step give 22.0\%, against 43.1\% for the two most confident ($p=6\times10^{-14}$; HumanEval: 28.7\% against 30.5\%, $p=0.76$), as PARD found for a fixed leftmost policy~\citep{hou2026pard}. \fo{}, which keeps the confidence check, reaches 82.3\% on GSM8K at $\B=256$ with 2.7 TPF in our own pipeline (prompt B).

\subsection{Distance cap}
\label{app:distcap}

The cap policy first commits what \fo{} would commit: the confident positions in a row, starting at the frontier. It then also commits each confident masked position $j$ with $j-f\leq k$, where $f$ is the leftmost position still masked after that. If no position qualifies, the fallback picks only among the masked positions within $k$ of the frontier. We decode greedily at the wide windows of \cref{tab:main} (\cref{tab:distcap}).

Caps of one or two positions keep \fo{}'s accuracy in six of nine model--task settings. How far ahead is safe depends on the model: Fast-dLLM v2 loses accuracy within a few positions, and LLaDA only beyond 32. A larger cap also adds little speed: TPF rises by at most 18\%, and in tokens per second the caps we timed ($k=2$ and 8) are 2--12\% slower than \fo{} at batch sizes 1, 8 and 32 (NLD-3B, $\B=128$, setup of \cref{tab:steady}).

\begin{table*}[tp]
\caption{\textbf{Distance cap:} accuracy (\%), its difference to $k=0$ ($\Delta$, points, $\pm$ half the width of its paired 95\% problem-bootstrap interval) and tokens per forward when confident commits and the fallback may reach at most $k$ positions past the frontier; $k=0$ is \fo{}, $k=\infty$ \off{}.}
\label{tab:distcap}
\centering
\scriptsize
\setlength{\tabcolsep}{2.2pt}
\begin{tabular}{llcccccccc}
\toprule
Model, task & & $k=0$ & 1 & 2 & 4 & 8 & 16 & 32 & $\infty$ \\
\midrule
NLD-3B, GSM8K & acc. & 87.6 & 87.9 & 87.9 & 87.8 & 87.6 & 87.1 & 86.4 & 79.1 \\
 & $\Delta$ & 0 & +0.4 & +0.4 & +0.2 & +0.1 & $-$0.5 & $-$1.1 & $-$8.5 \\
 & &  & {\tiny $\pm$1.3} & {\tiny $\pm$1.4} & {\tiny $\pm$1.5} & {\tiny $\pm$1.4} & {\tiny $\pm$1.4} & {\tiny $\pm$1.5} & {\tiny $\pm$2.0} \\
 & TPF & 2.77 & 2.81 & 2.82 & 2.85 & 2.87 & 2.86 & 2.80 & 2.27 \\[2pt]
NLD-3B, MATH-500 & acc. & 73.2 & 73.2 & 73.6 & 72.4 & 72.0 & 69.8 & 69.2 & 69.6 \\
 & $\Delta$ & 0 & +0.0 & +0.4 & $-$0.8 & $-$1.2 & $-$3.4 & $-$4.0 & $-$3.6 \\
 & &  & {\tiny $\pm$3.2} & {\tiny $\pm$3.6} & {\tiny $\pm$3.3} & {\tiny $\pm$3.5} & {\tiny $\pm$3.4} & {\tiny $\pm$3.5} & {\tiny $\pm$3.3} \\
 & TPF & 3.38 & 3.63 & 3.70 & 3.99 & 3.66 & 3.82 & 3.63 & 3.54 \\[2pt]
NLD-3B, HumanEval & acc. & 81.1 & 75.6 & 76.8 & 69.5 & 67.7 & 68.9 & 69.5 & 64.6 \\
 & $\Delta$ & 0 & $-$5.5 & $-$4.3 & $-$11.6 & $-$13.4 & $-$12.2 & $-$11.6 & $-$16.5 \\
 & &  & {\tiny $\pm$4.6} & {\tiny $\pm$5.2} & {\tiny $\pm$7.0} & {\tiny $\pm$6.7} & {\tiny $\pm$6.1} & {\tiny $\pm$5.8} & {\tiny $\pm$6.4} \\
 & TPF & 2.21 & 2.22 & 2.19 & 2.22 & 2.23 & 2.22 & 2.18 & 1.98 \\[2pt]
NLD-3B, MBPP & acc. & 72.5 & 73.3 & 71.2 & 73.3 & 72.0 & 69.8 & 71.2 & 63.5 \\
 & $\Delta$ & 0 & +0.8 & $-$1.3 & +0.8 & $-$0.5 & $-$2.6 & $-$1.3 & $-$9.0 \\
 & &  & {\tiny $\pm$3.3} & {\tiny $\pm$3.7} & {\tiny $\pm$3.7} & {\tiny $\pm$4.4} & {\tiny $\pm$4.4} & {\tiny $\pm$4.2} & {\tiny $\pm$4.8} \\
 & TPF & 2.20 & 2.23 & 2.11 & 2.04 & 2.14 & 2.12 & 2.00 & 1.82 \\[2pt]
NLD-8B, GSM8K & acc. & 90.1 & 91.6 & 91.3 & 92.0 & 91.7 & 91.4 & 90.5 & 87.3 \\
 & $\Delta$ & 0 & +1.4 & +1.1 & +1.9 & +1.6 & +1.2 & +0.4 & $-$2.9 \\
 & &  & {\tiny $\pm$1.2} & {\tiny $\pm$1.2} & {\tiny $\pm$1.3} & {\tiny $\pm$1.3} & {\tiny $\pm$1.4} & {\tiny $\pm$1.4} & {\tiny $\pm$1.8} \\
 & TPF & 3.78 & 3.61 & 3.67 & 3.45 & 3.63 & 3.60 & 3.59 & 3.37 \\[2pt]
NLD-8B, MATH-500 & acc. & 85.0 & 81.8 & 84.4 & 83.0 & 82.2 & 80.6 & 80.0 & 81.2 \\
 & $\Delta$ & 0 & $-$3.2 & $-$0.6 & $-$2.0 & $-$2.8 & $-$4.4 & $-$5.0 & $-$3.8 \\
 & &  & {\tiny $\pm$3.1} & {\tiny $\pm$3.0} & {\tiny $\pm$3.0} & {\tiny $\pm$2.9} & {\tiny $\pm$3.1} & {\tiny $\pm$3.3} & {\tiny $\pm$3.0} \\
 & TPF & 3.40 & 3.72 & 3.78 & 3.83 & 3.75 & 3.66 & 3.70 & 3.63 \\[2pt]
LLaDA (B), GSM8K & acc. & 82.3 & 82.7 & 82.1 & 82.0 & 81.0 & 80.6 & 80.8 & 35.5 \\
 & $\Delta$ & 0 & +0.4 & $-$0.2 & $-$0.3 & $-$1.4 & $-$1.7 & $-$1.5 & $-$46.9 \\
 & &  & {\tiny $\pm$1.7} & {\tiny $\pm$1.9} & {\tiny $\pm$1.9} & {\tiny $\pm$2.0} & {\tiny $\pm$2.0} & {\tiny $\pm$2.0} & {\tiny $\pm$3.0} \\
 & TPF & 2.72 & 2.78 & 2.81 & 2.86 & 2.92 & 2.97 & 2.97 & 1.76 \\[2pt]
Fast-dLLM v2, GSM8K & acc. & 85.4 & 84.5 & 85.2 & 83.5 & 81.4 & 79.5 & 79.6 & 79.4 \\
 & $\Delta$ & 0 & $-$0.9 & $-$0.2 & $-$1.9 & $-$4.0 & $-$6.0 & $-$5.8 & $-$6.1 \\
 & &  & {\tiny $\pm$1.7} & {\tiny $\pm$1.7} & {\tiny $\pm$1.9} & {\tiny $\pm$2.1} & {\tiny $\pm$2.1} & {\tiny $\pm$2.1} & {\tiny $\pm$2.3} \\
 & TPF & 3.14 & 3.22 & 3.22 & 3.26 & 3.27 & 3.29 & 3.31 & 3.32 \\[2pt]
Fast-dLLM v2, MATH-500 & acc. & 62.0 & 57.0 & 57.0 & 55.2 & 52.6 & 51.0 & 51.8 & 52.2 \\
 & $\Delta$ & 0 & $-$5.0 & $-$5.0 & $-$6.8 & $-$9.4 & $-$11.0 & $-$10.2 & $-$9.8 \\
 & &  & {\tiny $\pm$3.5} & {\tiny $\pm$3.7} & {\tiny $\pm$4.0} & {\tiny $\pm$4.1} & {\tiny $\pm$4.1} & {\tiny $\pm$3.9} & {\tiny $\pm$3.8} \\
 & TPF & 3.27 & 3.26 & 3.33 & 3.35 & 3.28 & 3.46 & 3.40 & 3.41 \\[2pt]
\bottomrule
\end{tabular}
\end{table*}

\subsection{Fast-dLLM v2 window sweep}
\label{app:fdllm}

Fast-dLLM v2 commits only inside 8-token sub-blocks of its 32-token training block, which limits how far ahead it can commit. Under \off{}, widening the window lowers accuracy step by step (\cref{tab:grid}): from the default sub-block 8 to $\B=128$, GSM8K drops from 84.8 to 79.4 and MATH-500 from 60.4 to 52.2 (Holm $p\leq0.002$), and most of the drop is already there at the whole 32-token training block, while \fo{} stays at 85.4--86.7 and 62.0--66.4. On code, \off{} falls from 63.4 to 50.0 on HumanEval and from 64.6 to 57.1 on MBPP at window 128 (Holm $p\leq0.003$), while \fo{} keeps 66.5 and 68.8 at the same TPF.

Inside the default decoder, \fo{} gains 6.0 points on MATH-500 (Holm $p=0.018$) and 7.1 on MBPP, as PARD reports for this model~\citep{hou2026pard}. Strict left-to-right decoding (sub-block 1) reaches 87.1\% on GSM8K and 67.2\% on MATH-500, and \fo{} in the default decoder comes within 0.4 and 0.8 points of it at 2.8 times the TPF. The Fast-dLLM v2 paper already noted that wider sub-blocks hurt its 1.5B model even at one token per step~\citep{wu2025fastdllmv2}.

\section{Roots and replays}
\label{app:roots}

\subsection{Roots}
\label{app:risk}

\begin{figure*}[tp]
 \centering
 \includegraphics[width=\textwidth]{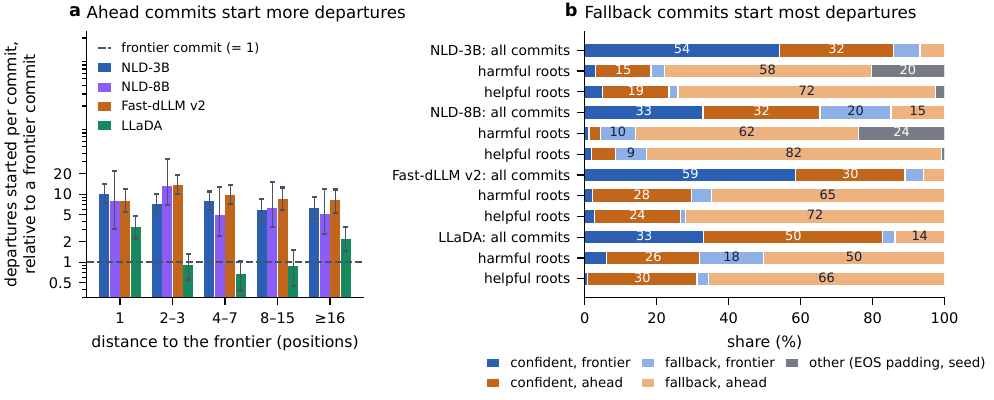}
 \vspace{-1.6em}
 \caption{\textbf{Fallback commits start most departures, and on the block-diffusion models, distance from the frontier adds risk at the same confidence.} \textbf{(a)} Departures (harmful or helpful) started per confident commit, by distance to the frontier, relative to a frontier commit at the same confidence (Poisson model with answer position, EOS token and task; log scale; 95\% intervals). \textbf{(b)} Shares of all commits, of harmful roots and of helpful roots by kind, at wide windows (NLD $\B=128$, Fast-dLLM v2 $\B=128$, LLaDA $\B=256$). Reference: each model's own \ltr{} decode; commits counted up to the root's step.}
\label{fig:risk}
\end{figure*}

\noindent \textbf{Definition.} The reference is the model's own \ltr{} decode at the same window (on Fast-dLLM v2, sub-block 1). Both decodes are greedy with the same prompt, so they agree token for token up to the first block that holds a differing answer position, and inside that block the root is the differing position that \off{} committed first. Distances are measured after each step: a token committed in a row with the frontier in the same step has $d=0$, like the frontier itself, so only commits with a masked position still to their left count as ahead. Inside the root block, we count only the commits made up to the root's step; the later commits are mostly at the frontier, and counting them would make frontier commits look safer than they are. Roots in the EOS padding after an \off{} answer that ends early are a separate kind; they are 20\% of the harmful roots on NLD-3B and 19\% on NLD-8B.

\noindent \textbf{Poisson model.} \Cref{fig:risk} shows the rate ratios of confident commits by distance (a) and the shares of commits and roots by kind (b); most fallback roots lie ahead of the frontier. For the model, we put every diffusion commit in a cell of task $\times$ distance bin (0, 1, 2--3, 4--7, 8--15, $\geq16$) $\times$ confidence bin (nine bins from $\geq0.999$ to $<0.50$, four of them below the threshold) $\times$ answer position (0--63, 64--255, $\geq256$) $\times$ EOS token. Per model, we fit $\log \mathbb{E}[\text{roots}] = \log(\text{commits}) + \beta_0 + \beta_{\text{task}} + \beta_{\text{dist}} + \beta_{\text{conf}} + \beta_{\text{pos}} + \beta_{\text{EOS}}$ to all roots, to harmful roots (with the two top confidence bins merged, since harmful roots at $\geq0.999$ are rare) and to confident commits only (\cref{tab:hazard}).

\noindent \textbf{Confidence and distance.} Confidence matters most: fallback bins start departures 90--1500 times as often as the $\geq0.999$ bin, and the position and EOS terms are small. At the same confidence, ahead commits still start harmful departures 4.0--15.2 times as often as frontier commits on the block-diffusion models, and 0.4--0.9 times as often on LLaDA, whose far-ahead predictions nevertheless disagree more often with the text written there later (\emph{Early predictions} below).

\begin{table*}[tp]
\caption{\textbf{Poisson model (reference: each model's own \ltr{} decode with the same prompt):} rate ratios of roots per commit against a frontier commit at the same confidence (nine bins), answer position, EOS token and task; wide windows (NLD $\B=128$, four tasks; Fast-dLLM v2 $\B=128$, GSM8K and MATH-500; LLaDA $\B=256$, prompts A and B); 95\% intervals from 500 problem resamples within tasks, given as the distance to their upper (superscript) and lower (subscript) end, or $\pm$ when the two are equal. Per-task values in the order GSM8K, MATH-500, HumanEval, MBPP (LLaDA: prompts A, B).}
\label{tab:hazard}
\centering
\scriptsize
\setlength{\tabcolsep}{4pt}
\renewcommand{\arraystretch}{1.3}
\begin{tabular}{lcccc}
\toprule
Term & NLD-3B & NLD-8B & Fast-dLLM v2 & LLaDA \\
\midrule
\multicolumn{5}{l}{\emph{all departures}} \\
distance 1 & 10.9$^{+2.0}_{-1.5}$ & 9.0$^{+1.2}_{-1.0}$ & 8.7$^{+1.6}_{-1.4}$ & 1.6$^{+0.4}_{-0.3}$ \\
distance 2--3 & 12.7$^{+2.2}_{-1.9}$ & 10.7$^{+1.7}_{-1.2}$ & 11.9$^{+1.8}_{-1.9}$ & 1.9$^{+0.2}_{-0.3}$ \\
distance 4--7 & 14.6$^{+2.9}_{-2.2}$ & 11.9$^{+2.0}_{-1.5}$ & 12.3$^{+2.4}_{-1.9}$ & 1.3 $\pm$ 0.2 \\
distance 8--15 & 10.2$^{+2.4}_{-2.0}$ & 14.3$^{+2.4}_{-2.1}$ & 10.8 $\pm$ 2.2 & 1.4$^{+0.3}_{-0.4}$ \\
distance $\geq16$ & 9.6$^{+2.6}_{-1.9}$ & 12.0$^{+2.7}_{-2.3}$ & 7.8$^{+2.0}_{-1.5}$ & 1.5$^{+0.3}_{-0.4}$ \\
\multicolumn{5}{l}{\emph{harmful departures}} \\
distance 1 & 9.5$^{+8.8}_{-3.6}$ & 7.7$^{+5.6}_{-3.2}$ & 11.4$^{+9.4}_{-4.7}$ & 0.9 $\pm$ 0.3 \\
distance 2--3 & 13.3$^{+11.0}_{-4.9}$ & 11.1$^{+7.9}_{-4.7}$ & 13.7$^{+11.0}_{-5.5}$ & 0.7 $\pm$ 0.2 \\
distance 4--7 & 15.2$^{+13.1}_{-6.3}$ & 10.3$^{+7.9}_{-5.2}$ & 14.7$^{+12.3}_{-6.1}$ & 0.5$^{+0.1}_{-0.2}$ \\
distance 8--15 & 9.6$^{+9.8}_{-4.5}$ & 14.6$^{+13.2}_{-7.2}$ & 11.2$^{+9.7}_{-5.5}$ & 0.4$^{+0.3}_{-0.2}$ \\
distance $\geq16$ & 8.0$^{+8.3}_{-4.2}$ & 4.0$^{+7.0}_{-4.0}$ & 9.4$^{+9.1}_{-4.5}$ & 0.9$^{+0.2}_{-0.3}$ \\
\multicolumn{5}{l}{\emph{confident commits only}} \\
distance 1 & 10.2$^{+4.1}_{-2.7}$ & 7.9$^{+13.8}_{-4.8}$ & 8.0$^{+3.9}_{-2.5}$ & 3.3$^{+1.5}_{-1.1}$ \\
distance 2--3 & 7.2$^{+2.8}_{-2.0}$ & 13.0$^{+19.8}_{-6.0}$ & 13.7$^{+5.3}_{-3.5}$ & 0.9$^{+0.4}_{-0.3}$ \\
distance 4--7 & 7.9$^{+3.1}_{-2.1}$ & 5.0$^{+7.9}_{-2.6}$ & 9.8$^{+3.8}_{-2.6}$ & 0.7 $\pm$ 0.3 \\
distance 8--15 & 5.8$^{+2.6}_{-1.7}$ & 6.3$^{+8.9}_{-3.0}$ & 8.6$^{+4.0}_{-2.7}$ & 0.9$^{+0.6}_{-0.5}$ \\
distance $\geq16$ & 6.2$^{+3.0}_{-1.9}$ & 5.2$^{+6.6}_{-2.6}$ & 8.2$^{+3.6}_{-2.8}$ & 2.2$^{+1.0}_{-0.7}$ \\
\multicolumn{5}{l}{\emph{ahead ($d\geq1$) vs frontier, within each task}} \\
per task & 12.3; 12.5; 17.0; 8.2 & 10.4; 10.8; 37.5; 9.2 & 11.0; 8.9 & 3.9; 0.8 \\
\bottomrule
\end{tabular}
\end{table*}

\begin{table*}[tp]
\caption{\textbf{Harmful and helpful roots at wide windows (the runs and reference of \cref{tab:risk}, commits counted up to the root).} Helpful: the \off{} answer is right and the reference answer wrong. Shares $\pm$ half the width of the 95\% interval; rate ratios (roots per commit, fallback vs confident) and the fallback share of the net accuracy gap to the reference (harmful minus helpful roots; a decomposition of the gap, not a causal estimate) with the distance to the upper (superscript) and lower (subscript) end of their 95\% interval, or $\pm$ when the two are equal; 500 problem resamples within tasks.}
\label{tab:mirror}
\centering
\scriptsize
\setlength{\tabcolsep}{3pt}
\renewcommand{\arraystretch}{1.3}
\begin{tabular}{lcccccccc}
\toprule
 & \multicolumn{2}{c}{roots} & fallback share & \multicolumn{2}{c}{fallback share of roots} & \multicolumn{2}{c}{fallback / confident per commit} & fallback share \\
\cmidrule(lr){2-3}\cmidrule(lr){5-6}\cmidrule(lr){7-8}
Model & harmful & helpful & of commits & harmful & helpful & harmful & helpful & of the net gap \\
\midrule
NLD-3B & 299 & 81 & 14\% & 61\% $\pm$ 6 & 74\% $\pm$ 9 & 20.2$^{+7.3}_{-5.7}$ & 19.1$^{+13.4}_{-6.7}$ & 56\% $\pm$ 8 \\
NLD-8B & 163 & 105 & 35\% & 72\% $\pm$ 7 & 90\% $\pm$ 5 & 31.5$^{+52.4}_{-14.0}$ & 19.9$^{+27.1}_{-8.1}$ & 38\%$^{+23}_{-69}$ \\
Fast-dLLM v2 & 239 & 75 & 11\% & 70\% $\pm$ 6 & 73\% $\pm$ 10 & 19.1$^{+6.8}_{-4.3}$ & 22.3$^{+14.2}_{-8.9}$ & 69\%$^{+9}_{-10}$ \\
LLaDA & 894 & 125 & 17\% & 68\% $\pm$ 3 & 69\% $\pm$ 8 & 10.3$^{+1.5}_{-1.4}$ & 10.7$^{+5.7}_{-3.1}$ & 68\% $\pm$ 4 \\
\bottomrule
\end{tabular}
\end{table*}

\noindent \textbf{Other references.} Two other references give the same result. With NLD's AR mode as the reference, fallback commits are 14\% and 41\% of the commits of NLD-3B and NLD-8B but start 65\% and 88\% of their harmful departures, and ahead commits start 1.6--3.6 times as many departures as frontier commits at the same confidence. With \seq{} as the reference, ahead commits start 2.2--19.5 times as many roots per commit as frontier commits, over 24 settings.

\noindent \textbf{Helpful departures.} A helpful departure, where the \off{} answer is right and the reference wrong, has a root defined the same way (\cref{tab:mirror}). Fallback commits start 69--90\% of the helpful roots, against 61--72\% of the harmful ones, and per commit they start helpful departures 11--22 times as often as confident commits do: the fallback is where decoding leaves the left-to-right path in either direction, and a guess is sometimes right. EOS padding roots are almost only harmful (NLD-3B: 20\% of harmful and 2\% of helpful roots).

\noindent \textbf{Net share of the gap.} Harmful roots outnumber helpful ones 1.6 to 7 times, and net of the helpful ones, departures that start at fallback commits make up 56--69\% of the accuracy gap to the reference on NLD-3B, Fast-dLLM v2 and LLaDA. On LLaDA, the share counts only failures with a root; counting its 127 cut-short failures, which have none, lowers it to about 58\%. On NLD-8B, where \off{} loses few answers, the interval includes zero. This share says where departures start; the replays of \cref{app:defer} test whether changing that step changes the answer.

\noindent \textbf{Which policy repairs which roots.} On the harmful problems of the wide-window runs, with NLD's AR mode as the reference on NLD, \eosf{} solves 82--96\% of the problems whose answer is cut short or whose root is EOS padding, but only 3--37\% of the others. \fo{} solves 70--100\% of them for every kind of root, on all four models.

\noindent \textbf{Narrow windows.} Fast-dLLM v2 at its default window (sub-block 8) and with the whole 32-token block shows the same pattern. Fallback commits are 12\% of all commits but start 69\% of the harmful departures, and confident commits at distance 1 start departures 33 times as often as frontier commits.

\noindent \textbf{Early predictions.} Roots depend on the reference decode. A second measure does not: whether a prediction made ahead of the frontier matches the token the model writes there later. In the left-to-right decodes of \cref{app:prop}, we take at every step the positions that \off{} would commit from the same forward pass, without committing them. For each position, we keep the first step at which \off{} would commit it, and call the prediction wrong if its token differs from the one the model commits when the frontier reaches it. Confident predictions in a row with the frontier are wrong 0.1--0.5\% of the time on all four models, confident ahead predictions 6.7--47\% and fallback ahead predictions 8--32\% (NLD and Fast-dLLM v2 at $\B=128$, four tasks on NLD; LLaDA has only 39 fallback predictions ahead of the frontier). In a logistic model with confidence bins, answer position and task, the odds of a wrong prediction grow with distance on every model, LLaDA included: 3.6--20 times the odds at $d=0$ for distance 1, and 71--4157 times for distance 16 or more. This measure covers only positions inside the answer. Most far-ahead commits in the \off{} decodes are EOS padding, which agrees with the reference and lowers the root rates at large distances (\cref{fig:risk}a).

\subsection{Replays}
\label{app:defer}

\begin{table}[!htb]
\caption{\textbf{At the root's step, committing only at the frontier repairs more failing answers than at matched control steps.} Repaired share (\%) of failing answers when the root's step or a matched earlier control step (ctrl.) is changed, and their paired difference (points, $\pm$ half the width of the 95\% problem-bootstrap interval); $n$: failing answers replayed. \Cref{tab:defer_sampled} gives sampled steps and helpful roots.}
\label{tab:defer_main}
\centering
\scriptsize
\setlength{\tabcolsep}{3pt}
\begin{tabular}{lllcccc}
\toprule
\textbf{Model} & \textbf{Task} & $\B$ & $n$ & \textbf{Root} & \textbf{Ctrl.} & \hd{\textbf{Root $-$}}{\textbf{ctrl.}} \\
\midrule
NLD-3B & GSM8K & 128 & 77 & 44.2 & 29.2 & $+14.9\pm9.0$ \\
NLD-3B & GSM8K & 256 & 325 & 29.5 & 22.3 & $+7.2\pm3.2$ \\
NLD-3B & MATH & 128 & 33 & 24.2 & 6.3 & $+17.9\pm14.9$ \\
Fast-dLLM v2 & GSM8K & 128 & 114 & 47.4 & 8.2 & $+39.2\pm8.7$ \\
Fast-dLLM v2 & MATH & 128 & 73 & 28.8 & 10.6 & $+18.2\pm8.7$ \\
\bottomrule
\end{tabular}
\end{table}

\noindent \textbf{Design.} For each problem, \off{} decodes once at batch size 1, and we choose the steps to change from this decode. A replay repeats the decode with one change at one step: at that step, the decoder commits only at the frontier, as \fo{} would. For a fallback step, this commits the frontier instead of the fallback's pick. At batch size 1, the replay computes exactly what the original decode computed up to that step, so any difference in the final answer comes from the change and from what the decoder does after it. With larger batches, bf16 results depend on which answers share a batch.

\noindent \textbf{Which steps we change.} \emph{Root}: the root's step, on harmful departures whose root is an ahead commit. \emph{Matched controls}: up to four earlier steps of the same answer. For each earlier step, we take its ahead commit that is most like the root, and rank the steps by four criteria in order: same kind as the root (fallback or confident; content or EOS padding), same block, fewest steps before the root, and closest distance. \emph{Sampled steps}: four steps per failing answer, drawn uniformly from all steps that commit ahead of the frontier, without using the root. Only afterwards do we label each sampled step as the root's step, before it or after it. \emph{Helpful roots}: on answers that \off{} gets right and the reference gets wrong, the root's step and two sampled steps.

\noindent \textbf{Selection.} We pick the root after we know the outcome, so the root's step could look special only because of how we choose it. The matched controls are still somewhat earlier than the roots (NLD-3B GSM8K, $\B=128$: step 9.3 against 12.6 within the block) and less often fallback (47\% against 65\%). The sampled steps do not use the root at all, and a logistic regression adjusts for the remaining differences: it predicts repair on all replayed steps of the failing answers from whether the step is the root's, where it lies relative to the root, its kind, distance, step, block and answer position, and the setting. The root effect, as an average marginal effect with a problem bootstrap, is $+7.3\pm3.1$ points on NLD-3B (555 answers) and $+30.8\pm7.7$ on Fast-dLLM v2 (239 answers).

\begin{table}[!htb]
\caption{\textbf{Replays at sampled steps and at helpful roots (\cref{app:defer}).} Sampled: steps drawn uniformly from all steps of the failing answer that commit ahead of the frontier, without using the root; repaired share (\%) at fallback and confident steps. Helpful: answers that \off{} gets right and the reference gets wrong; share (\%) broken by the same change at the root's step and, averaged, at two sampled steps of the same answers. $n$: sampled steps (four per failing answer) and helpful departures.}
\label{tab:defer_sampled}
\centering
\scriptsize
\setlength{\tabcolsep}{2.5pt}
\resizebox{\ifdim\width>\columnwidth\columnwidth\else\width\fi}{!}{\begin{tabular}{lcccccc}
\toprule
 & \multicolumn{3}{c}{sampled steps: repaired} & \multicolumn{3}{c}{helpful roots: broken} \\
\cmidrule(lr){2-4}\cmidrule(lr){5-7}
Setting & $n$ & fallback & confident & $n$ & root & sampled \\
\midrule
NLD-3B GSM8K, $\B=128$ & 440 & 29.6 & 17.3 & 19 & 36.8 & 7.9 \\
NLD-3B MATH-500, $\B=128$ & 172 & 4.2 & 3.0 & 14 & 35.7 & 17.9 \\
NLD-3B GSM8K, $\B=256$ & 1608 & 19.7 & 18.0 & 21 & 23.8 & 2.4 \\
Fast-dLLM v2 GSM8K, $\B=128$ & 592 & 16.6 & 9.6 & 50 & 34.0 & 5.0 \\
Fast-dLLM v2 MATH-500, $\B=128$ & 364 & 8.1 & 6.8 & 22 & 31.8 & 15.9 \\
\bottomrule
\end{tabular}}
\end{table}

\noindent \textbf{Results.} Against the matched controls, the root's step repairs more in all five settings (\cref{tab:defer_main}). On NLD-3B GSM8K, changing earlier steps also often repairs the failure, so there the failure is not tied to one step. Sampled steps (\cref{tab:defer_sampled}) repair 4--24\% of failing answers, and fallback steps repair more than confident ones in every setting. The two model families differ in where the useful steps lie. On NLD-3B GSM8K, sampled steps before the root repair 42\% ($\B=128$, 36 steps) and 27\% ($\B=256$, 303 steps) of the answers, against 22\% and 17\% after it, so these failures can often be undone before the first departure; on Fast-dLLM v2, steps before the root repair 6\% and 0\%, and the root's step matters most.

\noindent \textbf{Helpful roots.} On helpful departures, changing the root's step breaks the correct answer 24--37\% of the time, against 2--18\% at two sampled steps of the same answers: a paired difference of $+21$ to $+29$ points on the three GSM8K settings, with intervals above zero, and not significant on MATH-500. The root's step therefore matters in both directions, as the helpful roots of \cref{tab:mirror} suggest.

\section{DiffusionGemma}
\label{app:dg}

DiffusionGemma-26B~\citep{diffusiongemma2026}, a mixture-of-experts model with 3.8B active parameters, can re-sample accepted tokens, so it can undo an early guess. We test whether it uses this to fix ahead tokens. Its released sampler decodes 256-token answers block by block. Each step samples all positions, accepts the lowest-entropy set within an entropy bound of 0.1 and re-noises the rest with uniform random tokens, so an accepted token can still change later (temperature 0.8 to 0.4, at most 48 steps). We freeze accepted tokens at their first acceptance, and the same rule keeps choosing new positions, so the acceptance budget stays the same. We freeze either every accepted token, or only the tokens first accepted ahead of the frontier, at the frontier or at the first step. As a control, we freeze as many tokens at the same steps, chosen at random among the newly accepted ones. We run three seeds and average them within each problem (\cref{tab:dg}).

\begin{table}[!htb]
\caption{\textbf{DiffusionGemma-26B with frozen tokens.} Accepted tokens are frozen at their first acceptance, with the same acceptance budget. Three seeds averaged within each problem; four tasks pooled (2361 problems); HumanEval graded leniently. Differences to the default sampler $\pm$ half the width of the 95\% interval from resampling problems. Share of loss: each drop as a share of the 2.7-point drop from freezing all accepted tokens. The control freezes as many tokens at the same steps, at random positions.}
\label{tab:dg}
\centering
\scriptsize
\begin{tabular}{lccc}
\toprule
Frozen tokens & acc. & vs default & share of loss \\
\midrule
none (default) & 94.3 & -- & -- \\
frontier tokens & 93.8 & $-0.4\pm0.4$ & 15\% \\
step-0 tokens & 93.6 & $-0.7\pm0.4$ & 24\% \\
as many random (control) & 92.6 & $-1.6\pm0.5$ & 60\% \\
ahead tokens & 92.0 & $-2.2\pm0.5$ & 81\% \\
all accepted tokens & 91.5 & $-2.7\pm0.5$ & 100\% \\
\bottomrule
\end{tabular}
\end{table}

Freezing every accepted token costs 1.8--3.8 points across the three seeds, and tokens first accepted ahead of the frontier are revised 1.3--6.2 times as often as frontier tokens. Freezing only the ahead tokens reproduces 81\% of that loss, and freezing as many random tokens 60\%. Ahead tokens are 57--79\% of all tokens, so their share of the loss is about 1.1 times their share of the tokens; for frontier tokens, it is about 0.5 times.

\end{document}